\documentclass[acmsmall, natbib=false]{acmart}

\pdfoutput=1

\AtBeginDocument{%
  \providecommand\BibTeX{{%
    \normalfont B\kern-0.5em{\scshape i\kern-0.25em b}\kern-0.8em\TeX}}}

\usepackage{makecell}
\usepackage{dirtytalk}
\usepackage{graphicx}
\usepackage{amsmath}
\usepackage{multirow}
\usepackage{caption}
\usepackage[title]{appendix}

\usepackage[style=ACM-Reference-Format,backend=bibtex,sorting=none]{biblatex}
\setcopyright{acmlicensed}
\acmJournal{TECS}
\acmYear{2021} \acmVolume{1} \acmNumber{1} \acmArticle{1} \acmMonth{1} \acmPrice{15.00}\acmDOI{10.1145/3477026}

\author{Yeli Feng, Daniel Jun Xian Ng, Arvind Easwaran} 
\affiliation{%
  \institution{Nanyang Technological University }
  \streetaddress{50 Nanyang Ave, Singapore}
  \postcode{639798}
}
\email{ emails:  {yeli.feng, danielngjj, arvinde}@ntu.edu.sg }

\thanks{This article appears as part of the ESWEEK-TECS special issue and was presented in the International Conference on Embedded Software (EMSOFT), 2021. \\This research was funded in part by MoE, Singapore, Tier-2 grant number MOE2019-T2-2-040.}

\makeatletter \renewcommand{\fnum@figure}{Figure \thefigure} \makeatother

\begin{document}

\title{Improving Variational Autoencoder based Out-of-Distribution Detection for Embedded Real-time Applications}

\begin{abstract}
\vspace{2mm}
\hrule
\vspace{2mm}
Uncertainties in machine learning are a significant roadblock for its application in safety-critical cyber-physical systems (CPS). One source of uncertainty arises from distribution shifts in the input data between training and test scenarios. Detecting such distribution shifts in real-time is an emerging approach to address the challenge. The high dimensional input space in CPS applications involving imaging adds extra difficulty to the task. Generative learning models are widely adopted for the task, namely out-of-distribution (OoD) detection. To improve the state-of-the-art, we studied existing proposals from both machine learning and CPS fields. In the latter, safety monitoring in real-time for autonomous driving agents has been a focus. Exploiting the spatiotemporal correlation of motion in videos, we can robustly detect hazardous motion around autonomous driving agents. Inspired by the latest advances in the Variational Autoencoder (VAE) theory and practice, we tapped into the prior knowledge in data to further boost OoD detection’s robustness. Comparison studies over nuScenes and Synthia data sets show our methods significantly improve detection capabilities of OoD factors unique to driving scenarios, 42\% better than state-of-the-art approaches. Our model also generalized near-perfectly, 97\% better than the state-of-the-art across the real-world and simulation driving data sets experimented. Finally, we customized one proposed method into a twin-encoder model that can be deployed to resource limited embedded devices for real-time OoD detection. Its execution time was reduced over four times in low-precision 8-bit integer inference, while detection capability is comparable to its corresponding floating-point model. 
\end{abstract}


\maketitle
\hrule
\vspace{5mm}

\section{INTRODUCTION} \label{section:1}
Machine learning (ML) is rapidly finding its way into Cyber-Physical Systems (CPS). ML is highly data-driven. The reliability of its outputs decreases when test data deviates away from the distribution of training data (Out-of-Distribution problem - OoD). While ML adds new and attractive features to modern CPS, it also creates a new problem. Safety is critical for CPS because physical processes are involved, such as in autonomous vehicles, robotics, and many more. However, standard verification and validation methodologies in CPS design fail to handle the uncertainties in ML especially in applications where natural images create high input space dimensionality.  \cite{nguyen2015deep,eykholt2018robust}. 

In recent years, autonomous vehicles have caused several fatalities under rare or unbeknownst vision circumstances to ML, such as unexpected maneuvers from surrounding cars and pedestrians pushing bikes across a road in darkness. Hence, developing methods that can analyze hazards in real-time for ML-enabled CPS is very important.  

So far, many OoD detection methodologies have been proposed and tested over benchmark data sets that are much simpler than road scenes in autonomous driving. Broadly, these methods can be grouped into two categories. One group works through a supervised learning framework. By tapping into the statistics of neural weights in well-trained deep neural classifiers, novel metric scores that estimate the trustworthiness of a classifiers' outputs were proposed by \cite{lakshminarayanan2016simple, lee2018simple, oberdiek2018classification}. OoD and adversarial samples would receive trust scores lower than in-distribution (ID) samples. 

More recently, an unsupervised generative model approach has been explored by utilizing the information bottleneck property of Variational Autoencoder (VAE) to learn a posterior distribution approximation $q(z \mid x)$ to infer the unknown true latent distribution $p(z)$ of a training data set. Equation \ref{eq:1} is the current widely adopted evidence lower bound (ELBO) objective formulated by \cite{kingma2013auto}, where $\beta=1$. The first term on the right hand side, likelihood, measures the reconstruction accuracy of input. The second term measures the distribution's discrepancy in latent space. The learned latent space is better disentangled when adding an adjustable hyperparameter $\beta$ to balance the two terms \cite{higgins2016beta}. The $\beta$ value has to be carefully chosen per training data.

%\vspace{-3mm}
\begin{equation}
\label{eq:1}
\log p (x) \geqslant  \mathbb{E}_{q(z \mid x)}\big[ \log p (x \mid z) \big]  - \beta D_{KL}(q(z \mid x) \parallel p(z)) 
\end{equation}

Typically, the posterior approximation resides in a low dimensional latent space $\mathbb{Z}$, hundreds to thousands of times lower than the corresponding input space $\mathbb{X}$. Given test samples $x$, the VAE decoder reconstructs inputs with high likelihoods if inputs are ID, and otherwise with low likelihoods. The encoder of the VAE produces data-driven latent variables. The discrepancy of these latent variables' distribution to the latent prior $p(z)$ is commonly measured by Kullback–Leibler (KL) divergence. Detection methods \cite{daxberger2019bayesian, sundar2020out} utilized either the distribution discrepancy in latent space or the likelihood in input space as an OoD measure. Along this line, generative adversarial networks have also been explored by \cite{liu2019generative, nitsch2020out} to detect adversarial samples, a particular type of OoD data. Meanwhile, using likelihood alone failed to detect OoD in a specific situation has been revealed by \cite{nalisnick2018deep}. 

\begin{figure}[h]
  \centering
  \includegraphics[width=1\linewidth]{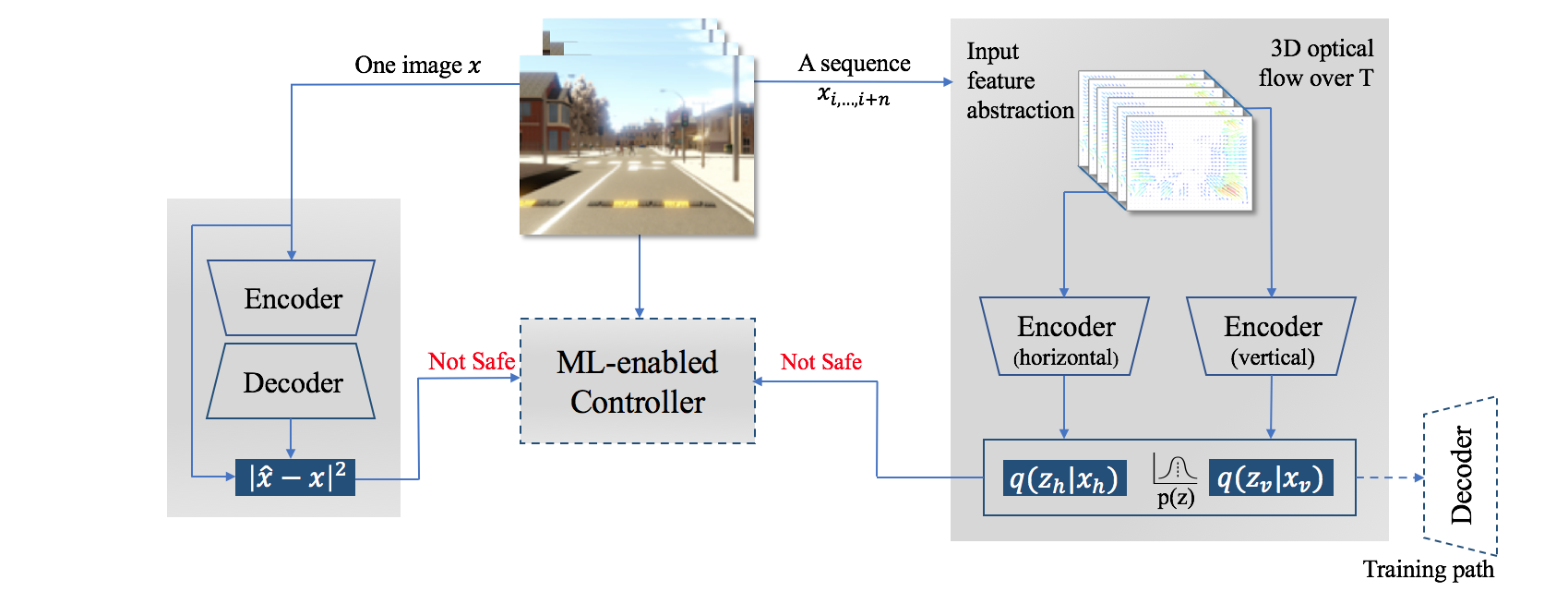}
  %\vspace{-6mm}
  \caption{A VAE-based OoD Detection Framework for ML-enabled Controllers }
  %\vspace{1mm}
  \begin{small}
  Our light-weighted spatiotemporal detection method conducts detection in the VAE latent space. It is depicted in the right-side grey box, where $z$ represents latent variables, $p(z)$ is unknown true prior distribution, and $q(z|x)$ is the data-driven approximation of posterior distribution. The left-side grey box shows how OoD is detected in the input space. To tackle the likelihood overlap concern, we propose an enhancement to the commonly used reconstruction error measure $|\hat{x} - x |^2$.
  \end{small}
  \label{fig:fig1}
\end{figure}

As of now, existing VAE-based OoD detection methods have reported good performance but only when training and test data are obtained from the same data set, and when detection were performed on standalone images instead of image sequences in video. Model generalization capability and robustness to noisy inputs have not been investigated in these works. Researchers have yet to discover effective methods that will perform robustly across multiple video data sets for real-time applications. Also, the latest advances in VAE formulation theory \cite{tolstikhin2017wasserstein} and its application for OoD detection \cite{nalisnick2018deep} have not been explored. With a focus on safety monitoring for ML-enabled autonomous driving, we probed into the existing likelihood-based detection methods, proposed a robust OoD detection architecture, and made significant enhancements. Contributions of our work are summarised as follows:  

\textbf{\textit{Light-weighted spatiotemporal detection}}: Hazardous motion around the ego vehicle is a unique and critical OoD factor in autonomous driving. We design a dual-input VAE to detect OoD motions in videos from the VAE latent space. There are two novelties in this design. First, it reduces the VAE input space complexity through feature abstraction by the means of 3D optical flow \cite{barron2005tutorial}. This makes it feasible to train a VAE with a Convolutional Neural Network (CNN) on high-dimensional time series data without a computationally costly Recurrent Neural Network. The other is the disentanglement of horizontal and vertical motions through the network architecture, shown in the right-side grey box in Figure ~\ref{fig:fig1}. Experiments in Section \ref{section:4.2.1} indicated that our unique design leads to robust performance, 42\% better than state-of-art approaches in F1 score.

\textbf{\textit{Optimal latent prior}}: 
The standard isotropic Gaussian is a common choice of the latent prior for OoD detection methods in the latent space $\mathbb{Z}$. Based on a novel autoencoder formulation proposed by \cite{tolstikhin2017wasserstein}, we devised an enhanced objective that trains a VAE model with the aggregated prior approximated from its training data set. This further improves performance of our light-weighted spatiotemporal detection method on Synthia \cite{bengar2019temporal}, a simulation driving data set, averagely by 8\% in F1 score. See evaluation in Section \ref{section:4.2.2}.

\textbf{\textit{Prior compensated OoD measure}}: 
Likelihood distributions of ID and OoD samples' can overlap when a VAE model is trained on a certain data set \cite{nalisnick2018deep}. Therefore, using likelihood alone as an OoD detection score is problematic under such conditions. Furthermore, this problem concerning training data is not symmetric. For example, a model trained with data set A can hardly detect data from an unrelated set B as OoD. Whereas training a model of the same network architecture but with data set B can detect set A data as OoD with high accuracy. See detailed illustration in  Appendix \ref{appendix:1}. To exploit this asymmetric behavior that was not captured by VAE, we leveraged the image complexity statistics in a model's training data set to formulate a complexity compensated likelihood score to improve OoD detection of visibility factors critical to driving safety. Over our implementation of related work \cite{cai:iccps2020}, the new score formula increases F1 score up to 8\% on fog and darkness OoD factors, as shown in Section \ref{section:4.5}.

\textbf{\textit{Robust model generalization capability}}:
Existing OoD detection methods in the CPS literature only reported performance on a hold-out partition from the same source of the training set. A robust OoD detector trained on data collected by driving through city A should not lose its prediction capability significantly when testing in city B. Performance evaluations across real-world and simulation driving data sets showed that our detection method generalized near-perfectly while related works failed badly. Average detection performance of our method drops trivially from  0.986 to 0.980 in AUROC\footnote{AUROC stands for Area Under the Receiver Operating Characteristic curve.}, 97\% better than the state-of-the-art approaches. See comparison study in Section \ref{section:4.3}.

\textbf{\textit{Designed for embedded real-time applications}}:
There is a lack of studies to evaluate whether the existing OoD detection methods are efficient for deployment to embedded devices for real-time applications. After compressed into an 8-bit integer version, the VAE model inference time was reduced over four times on Google Edge TPU. One OoD detection took 58.3 milliseconds on average on a Rasberry Pi mini-computer, while detection performance remains comparable to its corresponding floating-point model. See Section \ref{section:5.2}-\ref{section:5.3} for evaluation and comparison with related works.

In the rest of the paper, we first discuss related works in Section \ref{section:2} and subsequently describe our proposals in Section \ref{section:3}. Six groups of evaluation and comparison studies are presented in Section \ref{section:4}. Section \ref{section:5} is dedicated to deploying the proposed OoD detector and shows the related works to embedded devices and the experimental results.

\section{RELATED WORK}\label{section:2}
OoD detection for safety-critical ML-enabled autonomous CPS is an emerging field. There is a lack of closely related work in the literature. The mean absolute error between an image and its corresponding reconstruction from a VAE decoder was used as an OoD measure \cite{cai:iccps2020}. The KL-divergence values were utilized in $\beta$-VAE latent space to detect OoD samples \cite{sundar2020out, shreyas:emsoft2020wip}. In these works, VAE learns from standalone images. For real-time detection from video sequences, the inductive conformal prediction was leveraged by \cite{cai:iccps2020, shreyas:emsoft2020wip} to construct a detector that takes a sequence of OoD scores as input. These methods reported impressive performance in detecting OoD factors important to driving safety, such as rain intensities and illumination levels. However, whether these methods generalize well to a different driving data set have not been investigated in these works.

VAE-based OoD detection methods in input space have been studied extensively. Many works have focused on a particular type of domain shift, i.e., how well a model trained on image set A can detect images from set B as OoD. This type of OoD factor may not be desirable for driving safety. In a driving setting, it could be a requirement that a background shift from town to countryside shall not be taken as OoD. Nevertheless, the likelihood overlap problem pointed out by \cite{nalisnick2018deep}, and subsequent solutions in the literature inspire our work. Specifically, a VAE model trained on CIFAR10 \footnote{CIFAR10, SVHN, and MNIST series are widely used image data sets for machine learning and computer vision research. An explanation of likelihood overlap is provided in Appendix \ref{appendix:1}} images including common objects fails to detect most digit images from SVHN set as OoD, because the distribution of likelihood $\log p(\hat{x}_{SVHN} \mid M_{CIFAR10})$ overlaps with $\log p(\hat{x}_{CIFAR10} \mid M_{CIFAR10})$.    

The typicality set notion in \cite{choi2018waic} argued that likelihoods do not reveal where its probability mass is concentrated. By subtracting the likelihood's variance across independent samples from the posterior $q(z \mid x_{SVHN})$, the authors showed the problem can be solved well, with 16 samplings or 5 model ensembles to compute the likelihood variance.  

Based on a finding that the likelihood can be confounded by a general population level background statistics in genomics sequences, a likelihood ratio measure $\log p(\hat{x} \mid M) - \log p(\hat{x} \mid M_{b})$ was proposed by \cite{ren2019likelihood}, where model $M$ encodes genome information along with background information while model $M_b$ only encodes background statistics.  Their method also solves the likelihood overlap problem between the MNIST series image sets.

Extensive experiments with VAE and other popular generative models in \cite{serra2019input} suggest that this problem is due to the excessive influence of input complexity on likelihoods. The authors also proposed a likelihood ratio solution $\log p(\hat{x} \mid M)-\log p(\hat{x} \mid M_{o})$, where $M_{o}$ is a complexity estimation instead of a second VAE model as in \cite{ren2019likelihood}. In practice, the authors had chosen image compression ratios to estimate $\log p(\hat{x} \mid M_{o})$. 

For real-time CPS, the extra computing cost incurred by model ensembles or over a dozen independent samplings may render \cite{choi2018waic, ren2019likelihood} inefficient, especially on resource limited embedded CPS. Also, none of these likelihood correction proposals directly leverage known knowledge, e.g., ID statistics, in a training data set. 

OoD detection in latent space has only been recently explored. The encoder of a VAE discovers distributions over a  set of latent properties so that the decoder can reconstruct the input with a high likelihood. This approach requires the VAE latent space dimension to be much higher than the number of OoD factors. The dimensions and type of distribution of the unknown latent space are system parameters. So far, there is no mechanism to control which latent variable or variables respond to which OoD factor. To overcome this, a training set's KL-divergence scores were computed for each latent variable in \cite{daxberger2019bayesian} and simply half of the latent space with the highest scores were taken as OoD. Through testing, a subset of latent variables was identified in \cite{ shreyas:emsoft2020wip}. This subset encodes the most information about the training set.  A specific latent variable that responds the most to an OoD factor of interest was subsequently sought from this subset. When training an identical VAE architecture over a different data set, we observed that a particular latent variable that was in the aforementioned most informative subset could fall out of the set. It casts a concern about the model's generalization ability across data sets.

In the existing OoD detection literature, there is a lack of works utilizing the temporal dimension available in video, investigating model generalization capability, and exploiting the prior knowledge in the underlying data but were absent in the VAE learned latent space. This paper aims to address these missing points.

\section{PROPOSALS} \label{section:3}

In existing methods \cite{sundar2020out, cai:iccps2020, shreyas:emsoft2020wip}, VAE learns image sequences in videos as time-independent data points to reconstruct semantic details in underlying images or encode semantic properties in the latent space. If we consider that the most critical factor to road safety is always maintaining a safe margin of space around an autonomous vehicle, the robustness of these OoD detection methods are questionable. Firstly, semantic novelties do not principally explain for unseen motion. For example, instead of learning the spatial change of raindrops in an image sequence, existing methods rely on detecting unseen texture patterns from a single image.  Secondly, novel background semantics could trigger many OoD signals, mostly considered as false alarms which pose no risk to driving safety. This background semantics play a similar role to the genomics background in \cite{ren2019likelihood}, where it was compensated by a second model $M_b$ trained only on the background information. The inductive conformal prediction was applied by \cite{cai:iccps2020, shreyas:emsoft2020wip} to sequential OoD scores produced by VAE. This post-processing approach increased detection reliability but does not address the motion factor principally.

\subsection{Motion detection in spatiotemporal domain} \label{section:3.1} 

Recurrent neural networks (RNN) are a common choice of architecture for learning time series data. In \cite{li2018anomaly}, a GAN-based anomaly detector was designed to learn inputs from sensors and actuators during the normal working conditions of a CPS. The sensors' and actuators' inputs are low dimensional. For high dimensional image video, an RNN needs to be huge based on its neuron count, rendering it less computationally feasible for real-time CPS on embedded systems.

Optical flow is a classical computer vision technique widely used to estimate objects' actual movement in the physical world by computing changes in the neighboring pixels' intensity over time. If we want robust OoD detection for unusual motion, optical flow is an ideal feature extraction mechanism due to its deterministic computational outcome. Let \(I(x,t)\) denote pixel intensity in image frames and \(\nabla x\) be the unknown flow velocity in 2D space. The optical flow constraint equation $I(x+\nabla x,t+1)=I(x,t)$ assumes that the pixel intensity in a local region remains constant over a very short time frame. Solving the partial differential Equation \ref{eq:2} gives us the flow velocity. $n$ is the order of the Taylor series.

%\vspace{-0.5mm}
\begin{equation}
\label{eq:2}
\frac{\partial I}{\partial x}(x)+\frac{\partial I}{\partial x}(\nabla x) + \frac{\partial I}{\partial t}(t) + \frac{\partial^n I}{\partial x^n}(x)+\frac{\partial^n I}{\partial x^n}(\nabla x) + \frac{\partial^n I}{\partial t^n}(t)=0
\end{equation}

$\nabla x$ is a 2D vector field that represents surrounding object motions between consecutive image frames. A type of flow field example of driving on an open road is shown in Figure \ref{fig:fig1}, where far-away background semantics at the front view are filtered out, relative motions of the closer front road surface and lamp pole are transformed into motion vectors.  We found that anomaly event detection in video surveillance \cite{hu2004survey} to be broadly relevant to our work. However, unlike in the surveillance setting, where camera locations are fixed, motion in video feeds from a driving scene moves faster. The unique challenge here is to discriminate optical flow vectors caused by genuine OoD motions from those caused by continually changing backgrounds and other ID motion patterns in the environment. Today, the use of a deep neural network probably is the most effective technique to tackle this challenge . \cite{yu2021hmflow, costante2018ls}

Our motion OoD detection problem is different from anomaly detection for autonomous vehicles. Our problem deals with monitoring the potentially unreliable ML-enabled components in autonomous vehicles, specifically, the machine perception system. Anomaly detection in autonomous vehicles is not limited to the vision system (i.e., a extremely high data dimension problem). For example, non-vision low-dimension data from vehicle sensors are utilized in \cite{guo2019detecting} to to detect anomalies from pair-wise data correlation, such as between the acceleration and wheel torque. The approach proposed by \cite{ryan2020end} is framed in the end-to-end autonomous driving context. However, this approach detects for anomalies in vehicle control (steering, braking, and accelerating) through training a CNN model to learn the correlation between input driving video and human driving behavior. Although both approaches tackle the safety challenge in autonomous driving, our OoD detection problem focuses on the cause of uncertainty arising from machine vision. In contrast, anomaly detection focuses on the unexpected outputs from the vehicle controller.

To this end, we propose to extract ID motion series from image video into 3D optical flow tensors. While the spatial dimension of 3D optical flow remains the same as input images, the signal complexity is dramatically reduced. Therefore, it is feasible to train the time series with significantly fewer 3D convolutional network blocks. Let $\nabla x_{h,t}$ and $\nabla x_{v,t}$ be the 3D optical flow tensors for the horizontal and vertical directions respectively, the two 3D convolutional VAE encoders individually learn the input manifold, forming two separate latent sub-spaces. The latent sub-spaces are then concatenated as one input and then fed to the VAE decoder. The objective of our VAE network is to minimize the loss as given in Equation \ref{eq:3.1}, where $\phi$ and $\theta$ are model parameters of the encoder and the decoder respectively.  The second and third terms represent two encoders measuring the KL-divergence in the horizontal and vertical latent sub-spaces.  The first term refers to the decoder measuring the likelihood between encoders' input $\nabla x_{t}$ and the decoder's reconstruction $\nabla \hat{x_{t}}$ given input from the latent sub-spaces. A high level diagram of the network architecture is depicted in Figure \ref{fig:fig1}.

%\vspace{-1mm}
\begin{subequations} \label{eq:3}
\begin{equation}
\label{eq:3.1}
\begin{split}
\arg\min_{ \substack{\phi_h, \phi_v, \\ \theta;\nabla x_{t}}} & - \; \mathbb{E}_{q(z_{h,v} \mid (\nabla x_{t})}\big[ \log p (\nabla x_{t} \mid z_{h,v}) \big] \\
& + \; D_{KL} \big( q(z_{h} \mid \nabla x_{h,t}) \parallel p(z_{h}) \big) 
 + \; D_{KL} \big( q(z_{v} \mid \nabla x_{v,t}) \parallel p(z_{v}) \big)  
\end{split}
\end{equation}  

%\vspace{-1mm}
\begin{equation}
\label{eq:3.2}
\begin{split}
- \mathbb{E}_{q(z_{h,v} \mid (\nabla x_{t})}\big[ \log p (\nabla x_{t} \mid z_{h,v}) \big] & = 
\; \frac{1}{2} \parallel \nabla \hat{x_{t}} - \nabla x_{t} \parallel^{2} \; + \; M \ln(2\pi) \; = \; \frac{M}{2}MSE(\nabla \hat{x_{t}}, \nabla x_{t} ) + c
\end{split}
\end{equation}
\end{subequations}

Bernoulli distribution, as proposed in \cite{kingma2013auto}, is optimal for modeling black and white image data like MNIST because pixels value are either 0 or 1. Optical flow values and normalized color image pixel values are real numbers in a continuous domain. In practice, we adopt a common approach that modeling the training data with a Gaussian distribution with identity covariance. Hence a Gaussian decoder, where the negative log-likelihood term in Equation \ref{eq:3.1} is transformed into mean square error (MSE) in implementation, as shown in Equation \ref{eq:3.2}. See Appendix \ref{appendix:2} for the derivation steps. M is the input space dimension, a multiplication of length, width, and depth of an input. OoD score is a summation of $ \sum_{i=1}^{n} D_{KL}$ in horizontal and vertical latent sub-spaces. $n$ is the number of dimensions in the latent sub-space. Experiments with depth values of 6 and 1 showed that the proposed feature abstraction in spatiotemporal space leads to the robustness of OoD detection.

%\vspace{-2mm}
\begin{figure}[h]
  \centering
  \includegraphics[width=0.9\linewidth]{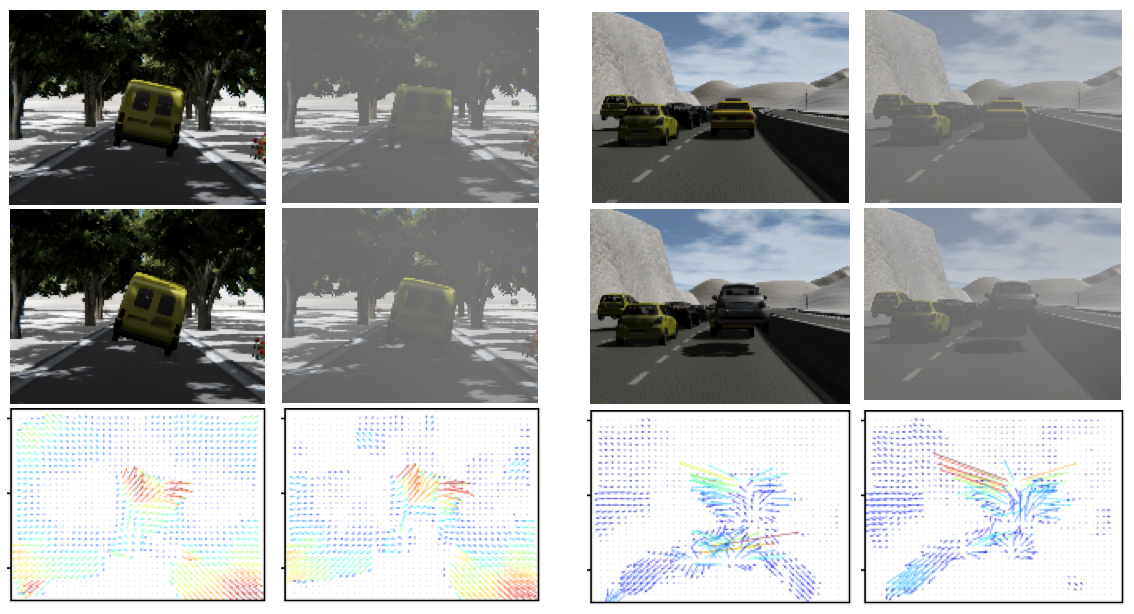}
  %\vspace{-1mm}
  \caption{ Examples of optical flow features extracted from image sequences including OoD factors. }
  \begin{small}On the left example, a car at the front flips over. On the right, a vehicle is being spawned onto the driving lane. \end{small}
  \label{fig:fig2}
\end{figure}

%\vspace{-4mm}
Can hazardous motion features still be robustly extracted when OoD factors that affect visibility, such as fog, co-occur? In Figure \ref{fig:fig2}, we visualized the flow vectors of two OoD motion examples. Flow vectors shown in the bottom row are computed from the pair of image frames from the row above. Clean image sequences were obtained from \cite{bengar2019temporal}. The fog was rendered over the clean sequences with dynamic effects through Autodesk Maya software. Both cases simulate that a car is driving ahead. Optical flow vectors in the front view are visualized as colored vectors. A cluster of red represents the prominent motion between two video frames. We can see that the fog almost has no impact in both cases.  
Qualitatively, we showed that a robust representation of motion around an autonomous vehicle can be extracted through optical flow under diverse visibility conditions. Since this representation is abstract, environmental background semantics will not be encoded into the latent space, which encourages model generalization and this better suits to driving environments that are different from the training data set. In related works \cite{ shreyas:emsoft2020wip, daxberger2019bayesian}, an informative subset of the latent space has to be experimentally selected for good OoD detection performance. A consequential benefit of this proposed method is that the entire latent space or a particular sub-space can be used for OoD detection.

\subsection{Optimal prior distribution in latent space} \label{section:3.2} 

Commonly, a simple prior $\mathcal{N}(0,\mathbf{I})$ is used to represent the true but unknown latent prior $p(z)$, and the KL-divergence is used to compute the distribution discrepancy between the encoder output and the prior, as given in Equation \ref{eq:4}. 

%\vspace{-4mm}
\begin{equation}\label{eq:4}
\begin{split}
D_{KL} \big( q(z \mid \nabla x_t) \parallel p(z)\big) & = - \frac{1}{2} \Bigg[ \frac{1}{\sigma_{p}^{2}}\mathbb{E}_{q} \bigg( (\nabla x_t - \mu_{p})^2 \bigg) - \log\bigg(\frac{\sigma_{q}}{\sigma_{p}} \bigg) -1 \Bigg] 
\end{split}
\end{equation}

For robust OoD detection in the latent space, it is critical that the posterior accurately approximates underlying ID data properties.  Intuitively, assuming that any underlying data follows a simple prior $\mathcal{N}(0,\mathbf{I})$ is not optimal. One way for improvement is to optimise the prior and not just the encoder and decoder \cite{hoffman2016elbo}. 

Substituting the simple prior with an approximation of aggregated posterior $ q(z) = \frac{1}{N}\sum_{n=1}^N q(z \mid x_n) $ was proposed by \cite{tomczak2018vae}. A Student's t-distribution prior allowed a more robust approximation of the underlying data in the presence of outliers \cite{abiri2020variational}. A Dirichlet prior was chosen by \cite{xiao2018dirichlet} to model a latent representation of categorical distribution, specifically the class labels distribution, for a downstream object classification task. Since the input and latent spaces in our VAE proposal encode the same physical meaning but because the latent dimension is dramatically reduced, we decided to identify an optimal prior from model's training data.

\begin{figure}[h]
  \centering
  \includegraphics[width=0.9\linewidth]{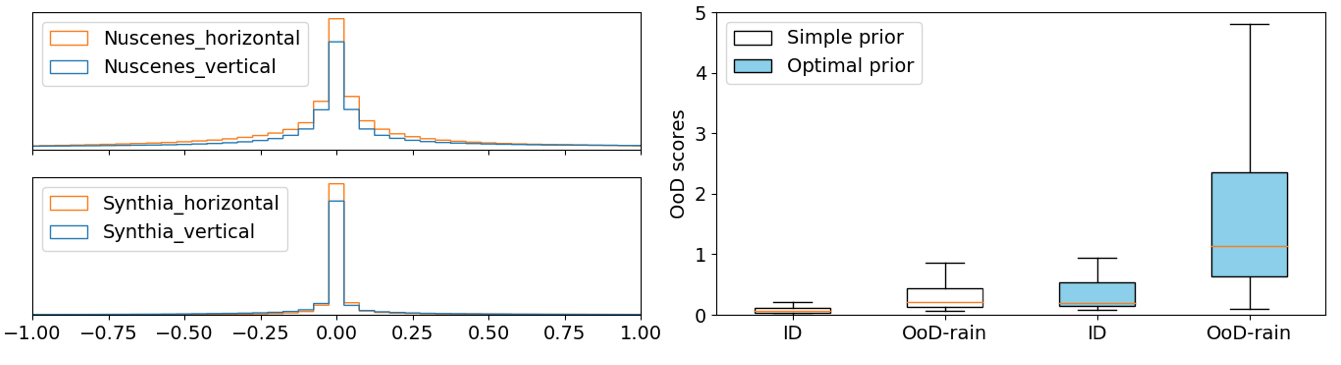}
  %\vspace{-6mm}
  \caption{ Optimal prior distribution $p(z)$.}
  %\vspace{1mm}
  \begin{small}The left column shows histograms of optical flow $ \nabla x $ from the underlying ID data. The right column shows box plots of OoD scores, $D_{W2}$ and $D_{KL}$, from the corresponding test set. \end{small}
  %\vspace{-3mm}
  \label{fig:fig3}
\end{figure}

We select a random subset from training data and plot a histogram of optical flow values $\nabla x$, as shown in the left column of Figure \ref{fig:fig3}. Distributions of flow values in both data sets approximate to Gaussian in horizontal and vertical directions but with very small deviations. In practice, we replace the variance of the simple prior $\mathcal{N}(0,\mathbf{I})$ with $\sigma_p$ approximated from a random partition of the training set $\nabla x$. 

%\vspace{-4mm}

Since the $\sigma_p$ value is very small here, training with Equation \ref{eq:4} as the objective will be unstable. To overcome this, we adopt a novel autoencoder formulation from \cite{tolstikhin2017wasserstein}, the Wasserstein Auto-Encoder (WAE). Motivated by the transportation theory in mathematics, the objective of WAE is to minimize the transportation cost between true data distribution $p(x)$ and a generative model, i.e., the decoder modeled after latent distribution $ q(z \mid x) $, denoted as $c(x,z)$ in Equation \ref{eq:5}. Notation $\mathcal{D}$ in the second term of the objective is a distance metric that encourages the encoded training distribution to match its prior $p(z)$. Both the cost and metric functions can be arbitrary, so WAE is a non-parametric probabilistic model. $\lambda>0$ is a hyperparameter. With WAE, we can use a metric other than KL-divergence as an OoD measure and any reconstruction error function to regularize the decoder. 

%\vspace{-2mm}
\begin{equation}
\label{eq:5}
\arg\min  \; \mathbb{E}_{p(x)} \mathbb{E}_{q(z \mid x)}\big[ c(x, z) \big]  \;+\; \lambda D_{z} \big(q(z \mid x) \parallel p(z) \big)
%\vspace{0.5mm}
\end{equation}

With this new objective, we revise the objective in Equation \ref{eq:3} by substituting the $D_{KL}$ penalization with a 2-order Wasserstein metric $D_{W2}$, as given in Equation \ref{eq:6}. The decoder is still regularized by MSE loss and $\lambda=1$. For isotropic Gaussian, covariance matrix $C_p = \sigma_p^2 \mathbf{I}$ . A $D_{W2}$ penalization ensures the training is stable even when $\sigma_p$ is close to zero.

%\vspace{-2mm}
\begin{equation}
\label{eq:6}
D_{W2} = \; \parallel \mu_p - \mu_q \parallel_2^2 \; + \;  \big( C_p + C_q -2(\sqrt{C_q} C_p \sqrt{C_q} ) \big)
%\vspace{1mm}
\end{equation}

How will the optimal prior impact the OoD scores $D_{W2}$? Will the optimal prior lead to model overfitting? The box plots of $D_{KL}$ and $D_{W2}$ in Figure \ref{fig:fig3}'s right column show that the difference between mean OoD scores of ID and OoD rain test samples increases in the model trained with an optimal prior. In experiments, we observed a consistent detection performance enhancement over test samples from the same or from different data sets, which indicates that the proposed optimal prior does not cause overfitting. Refer to Section \ref{section:4.3} for a quantitative evaluation.

\subsection{Compensated OoD measure in input space} \label{section:3.3}

Section \ref{section:3.1} explained that the motion abstraction with optical low is robust under diverse visibility conditions. However, we can't rely to this abstraction when visibility is close to zero. To address such extreme visibility conditions, we propose an improvement to OoD detection in the input space.

In Section \ref{section:2}, we introduced the likelihood overlap problem and commented that most existing likelihood compensation approaches $\log p(\hat{x} \mid M)-c(x,\hat{x})$ are not suitable for real-time CPS application because of the extra computing cost incurred. There is an exception where authors of \cite{serra2019input}  proposed to compensate with an image complexity ratio estimated by compression algorithms such as JPEG2000. 

\begin{figure}[h]
  \centering
  \includegraphics[width=1\linewidth]{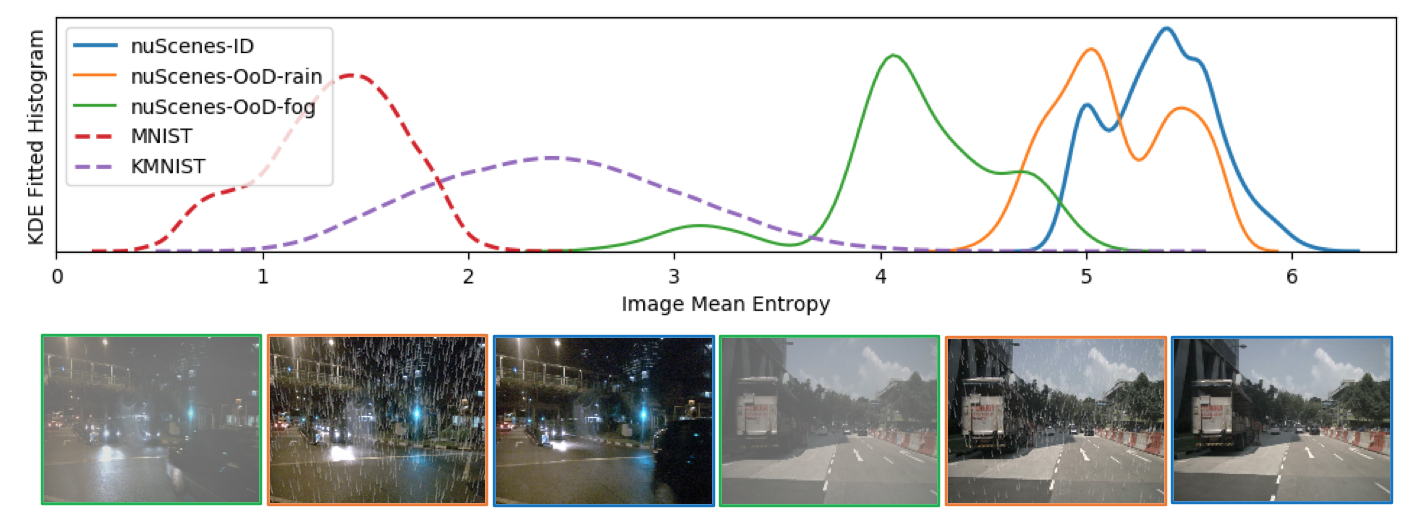}
  %\vspace{-4mm}
  \caption{ Complexity Compensated OoD Measure }
  %\vspace{1mm}
  \begin{small}The top row shows kernel density estimation (KDE) fitted histograms of the image mean entropy. The fog and rain OoD factors are rendered over ID images from the nuScenes data set with Autodesk Maya software's dynamic effects. Examples of OoD fog, OoD rain, and ID images are shown at the bottom. \end{small}
  \label{fig:fig4}
\end{figure}

Instead of compression ratio, we use image entropy as a complexity measure to study the likelihood overlap problem. Information entropy is a widely used complexity measure not limited to just images. 2D image entropy measures the complexity in local neighborhoods of an image. From the histograms of image entropy in Figure \ref{fig:fig4}, we can see that the fog factor does cause a considerable reduction of entropy while the rain factor almost has no impact. Also, images in the nuScenes data set are more complex than the benchmark data sets such as MNIST and KMNIST.

Inspired by related works, we propose to improve the detection of low visibility conditions by calibrating the likelihood with a data-driven baseline. The baseline is the mean entropy of model's training data, denoted as $entropy(M)$ in Equation \ref{eq:7}. $entropy(x)$ is the mean entropy of a test image. There does not exist ubiquitous background statistics in our problem, as in the genomics sequence study \cite{ren2019likelihood}. So the likelihood is compensated only when a test image's mean entropy is lower than the baseline. In experiments, we use $MSE(x, \hat{x})$ instead of likelihood $\log p(\hat{x} \mid M)$.

%\vspace{-2mm}
\begin{equation}
\label{eq:7}
score_{OoD} = \min \bigg [ 1,\; \frac{entropy(M)}{entropy(x)}  \bigg ] * \log p(\hat{x} \mid M) 
\end{equation}

How effective is this metric compensation proposal? In Appendix \ref{appendix:1}, we demonstrated that a model trained with KMNIST cannot differentiate MNIST images from KMNIST. Their likelihoods histogram almost overlap. Since most MNIST images can easily be separated from KMNIST images by entropy (see in Figure \ref{fig:fig4}), calibrating likelihoods improves cross data-domain OoD detection performance from 0.504 to 0.939 in AUROC. 

We shall note that this likelihood calibration is explicitly designed for robust detection of low visibility factors in an autonomous driving setting. Its effectiveness is evaluated in Section \ref{section:4.2} using visibility factors such as fog and low light.

%\vspace{2mm}
\section{EVALUATION} \label{section:4}

The evaluation part has seven sections. Section \ref{section:4.1} describes the data sets used and hte experiment setup. Sections \ref{section:4.2}-\ref{section:4.6} evaluate the OoD detection performance of proposed methods and compare them with related works. Section \ref{section:4.7} covers the computational cost. The feasibility and customization required for real-time OoD detection on off-the-shelf embedded devices is studied in Section \ref{section:5}.

\subsection{Experiment setup}\label{section:4.1}

We implemented two of the most closely related OoD detection methods, \cite{cai:iccps2020} and \cite{shreyas:emsoft2020wip} from the autonomous CPS domain, for performance comparison. The performance advantages of our light-weighted spatiotemporal detection method were demonstrated with input sequences of different lengths in time. Our improvement to the likelihood overlap issue was validated. Experiment details are summarized in the below. 

\textbf{\textit{OoD factors and Experiment Data Sets}}  We are interested in detecting factors that are potentially dangerous to safe driving. OoD factors evaluated were close hazardous motion, rain, fog, and low light. We selected a driving simulation data set Synthia  \cite{bengar2019temporal} and a real-world driving data set nuScenes mini \cite{caesar2020nuscenes} to partition into training and test sets. 

Synthia has 288 video scenes captured from a driving car in a virtual world at 25fps with a scene length between 10 to 30 seconds and a rich mixture of weather, landscape (i.e., city, town, and highway), and dynamic and static objects (car, pedestrian, bicycle, wheelchair, veneration, traffic light, etc.).  We manually curated 94 video scenes from Synthia to form a training set of 27519 images. The nuScenes mini has 10 driving video scenes recorded in city environment. The first 48 frames of each nuScenes mini video were kept for testing, and the remainder of 1870 images were used for model training. Both training sets only include ID data. 

\begin{figure}[h]
  \centering
  \includegraphics[width=1\linewidth]{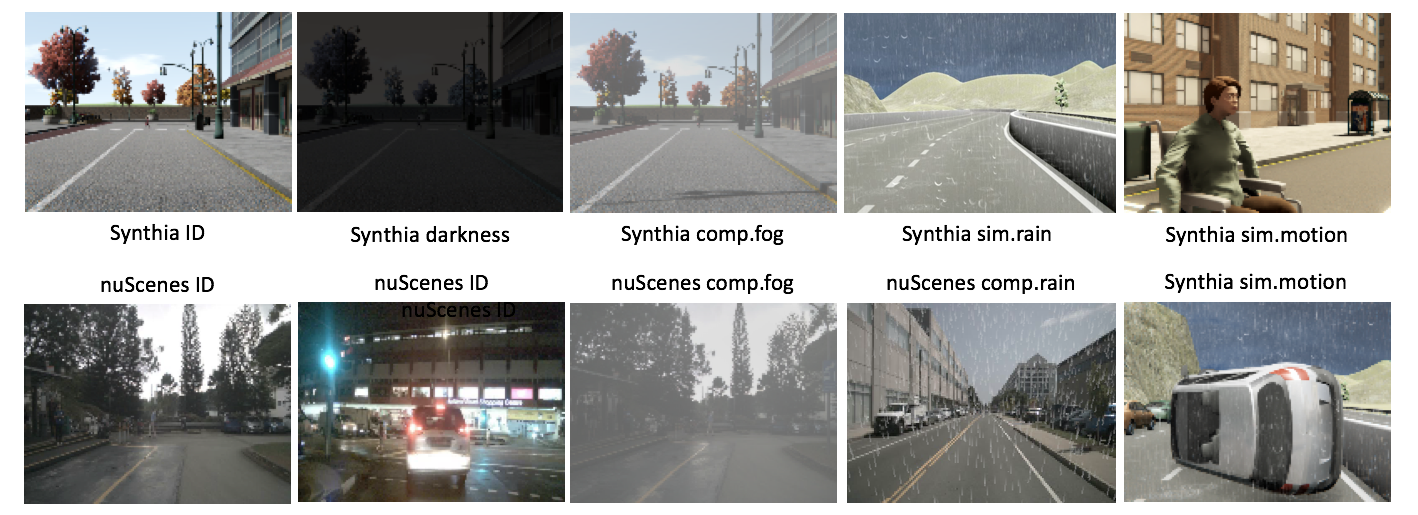}
  %\vspace{-4mm}
  \caption{ Examples of ID and OoD Images used in Evaluation}
  \label{fig:fig5}
\end{figure}

Details of test sets are summarized in Table \ref{table:1}, where the source and sample numbers of images or video episodes for each class are listed in the third column. OoD classes starting with \say{comp} were generated via rendering dynamic effects rain and fog over ID image sequences using Autodesk Maya software. Images in the darkness OoD class were generated by reducing ID images' brightness in the HSV color space. OoD classes starting with \say{sim} are original images from the Synthia data set. In the sim.motion OoD class, each episode is 60 frames long with at least one motion OoD occurrence. Some examples of ID and OoD images are shown in Figure \ref{fig:fig5}. 

\begin{table}
  \caption{ Test Sets and Definition of OoD} \label{table:1}
  %\vspace{-2mm}
  \begin{tabular}{l|l|l|l}  
    \hline
    Class & Group & Images & Description  \\
    \hline
    \small ID  & & \makecell[l] { \small Synthia (6660) \\ \\ \small nuScenes (480) }  & \makecell[l] { \small Synthia  (day time, dry weather, on straight \\ \small street or highway, stopping at cross junctions.)  \\ \small nuScenes (day or night, city street, \\ \small driving straight or fast turns.) } \\ \hline  
    \makecell[l] {\small OoD \\ \small sim.motion} & motion & \makecell[l] { \small Synthia \\ \small (60 episodes)} & \makecell[l] { \small One of the five situations occurred without or \\ \small with rain: abrupt cut of the driving lane by\\ \small spawned car or pedestrian;  car crash in front of \\ \small the driving car; inter-vehicle distance \\ \small drops drastically (i.e., emergency braking); \\ \small the driving car vibrates vertically; \\ \small the driving car turns at a road intersection. } \\ \hline        
    \small OoD sim.rain & \makecell[l]{ \small motion\\ \small visibility} & \small Synthia (7740) & \small Rain generated by Synthia 3D simulation \\  \hline    
    \small OoD comp.rain &  \makecell[l]{ \small motion\\ \small visibility} & \makecell[l] {\small Synthia (480) \\ \small nuScenes (480) } & \small 2D composition rain rendered over ID test set \\ \hline
    \small OoD comp.fog & visibility & \makecell[l] {\small Synthia (480) \\ \small nuScenes (480) } & \small  2D composition fog rendered over ID test set \\ \hline
    \small OoD darkness & visibility & \makecell[l] {\small Synthia (6660) } & \small Reduce brightness of ID test set images\\ \hline
\end{tabular}
%\vspace{-4mm}
\end{table}

\textbf{\textit{VAE networks}} 
To fairly compare the three detection methods, we used an identical CNN architecture for the encoders and decoders, because this network architecture was used by \cite{cai:iccps2020, shreyas:emsoft2020wip}. Their corresponding comparison studies will be carried out in Sections \ref{section:4.2}, \ref{section:4.3}, \ref{section:4.5} and \ref{section:4.7}. All input images were resized to 120x160 in pixels. The encoder has four CNN blocks of 32/64/128/256 x (5x5) filters with exponential linear unit (ELU) activation and BatchNorm. The decoder is architecturally symmetric. The extra layers in \cite{cai:iccps2020, shreyas:emsoft2020wip} networks are two MaxPool in the encoder and two MaxUnPool in the decoder, after the CNN blocks. Whereas our network does not have these pool layers as we use larger strides in the CNN blocks.

The dimension of latent space varies. Following the original proposals, the latent dimension is 1024 and 30 for \cite{cai:iccps2020} and \cite{shreyas:emsoft2020wip}, respectively. The latent dimension in our method is 12 for each sub-space. The depth of 3D optical flow is 6, computed from 7 consecutive image frames. In sections \ref{section:4.4} and \ref{section:4.6}, we show how different numbers of CNN layers, dimensions of latent space, and depth of input size influence the overall performance of the proposed OoD detection methods.

\textbf{\textit{Hyperparameters}} 
All VAE models were trained with the Adam optimiser using PyTorch. For the Synthia training set, model \cite{cai:iccps2020} was trained with constant learning rates \(10^{-4} \) and \(10^{-5} \) for 200 and 150 epochs, respectively. Model \cite{shreyas:emsoft2020wip} was trained with a constant learning rate \(10^{-5} \) for 100 epochs. These learning rates and training epochs were adopted from \cite{cai:iccps2020, shreyas:emsoft2020wip}. We tested a faster learning rate \(10^{-3} \), however it degraded the model performance in these methods. Our VAE model was trained with a constant learning rate \(10^{-4} \) for 100 epochs.

Increasing total training epochs did not enhance performance with the Synthia training set but was the opposite with the nuScenes-mini training set. After a quick search, all corresponding nuScenes models were trained with 600 epochs. 

In \cite{shreyas:emsoft2020wip} method, we set the $\beta$ value to 1.4 and selected the nine latent variables that encode the most information, as proposed in the original paper. See Appendix \ref{appendix:4} for the selection algorithm.

\textbf{\textit{Evaluation metrics}} 
Equation \ref{eq:8} defines the OoD calculation formula used to analyze experiments in this paper. It is a summation of distribution discrepancies between a test sample and the latent priors in horizontal and vertical sub-spaces. 
\begin{equation}
\label{eq:8}
Score = Score_{h} + Score_{v} = \sum_{i\in\mathbb{Z}_{h}}D_{h}(q(z_{h}| \nabla x_{h}) \mid p(z_{h})) + \sum_{j\in\mathbb{Z}_{v}}D_{v}(q(z_{v}|\nabla x_{v}) \mid p(z_{v}))
\end{equation} 

We used AUROC to compare the performances of OoD detection methods across different score thresholds. In true positive rate (TPR) 95\%, i.e., tolerating 5\% miss-out, we compared point performances of OoD detectors in precision and F1 score metrics. When TPR is fixed, precision, $TP / (TP+FP)$, reveals the false positive rate, while the F1 score conveys the balance between precision and sensitivity $TP / (TP+FN)$. 

\subsection{Robust Motion OoD Detection} \label{section:4.2}
In this section, we first compared the performances of three VAE-based detection methods over the motion OoD group. Next, the benefit of optimizing prior distribution was analysed over the motion OoD group. Finally, the extra performance benefit from detecting OoD in a latent sub-space was shown. For simplicity, our detection method proposed in Section \ref{section:3.1} is termed as \textit{bi3dof} and its enhanced version proposed in Section \ref{section:3.2} as \textit{bi3dof-optprior}. 

\begin{figure}
  \centering
  \includegraphics[width=1\linewidth]{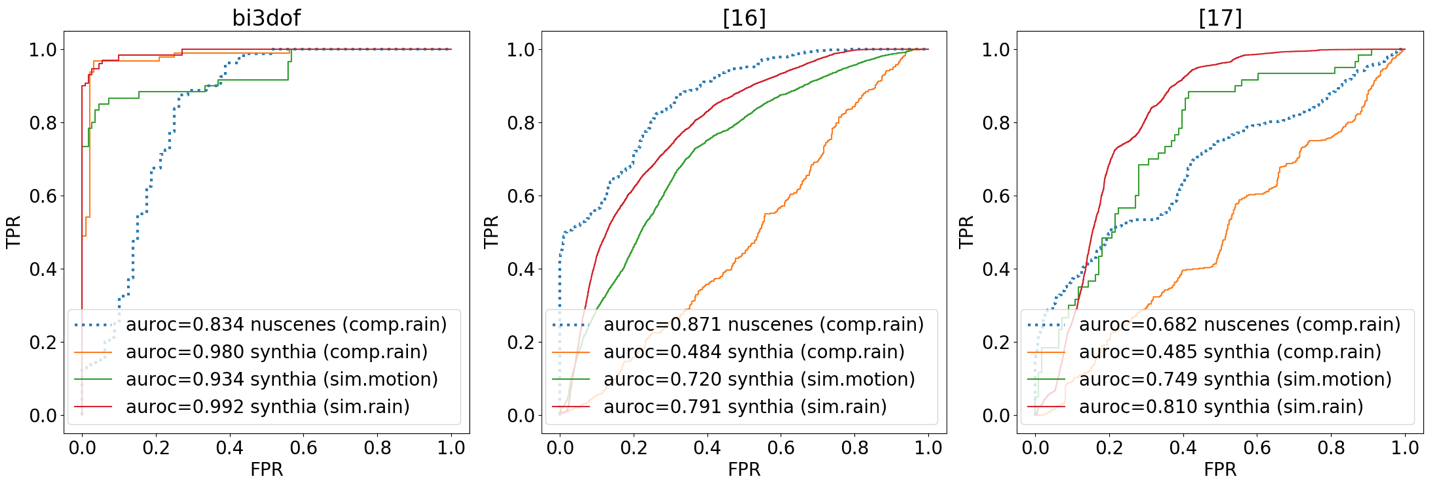}
  \vspace{-8mm}
  \caption{A Summary of the Performances of Detection Methods Over Motion OoD Factors}
  \label{fig:fig6}
  %\vspace{-4mm}
\end{figure}

\subsubsection{\textbf{Comparison Study}} \label{section:4.2.1}
Figure \ref{fig:fig6} summarizes each method's detection capacity broken down by OoD classes.  We can observe the bi3dof trained on the Synthia data set is very robust. Its overall detection performance varies between 0.980 to 0.992 across all motion OoD classes. Whereas, the best results from \cite{cai:iccps2020} and \cite{shreyas:emsoft2020wip} are 0.791 and 0.810 respectively on the Synthia data set.

Methods \cite{cai:iccps2020, shreyas:emsoft2020wip} failed to detect the comp.rain class on the Synthia data set, both AUROC values are bellow 0.5. But on the nuScenes test set, their detection performance over the comp.rain class increases, with \cite{cai:iccps2020} scores 0.871, higher than bi3dof which is 0.834.  The nuScenes training set size is 6.7\% of the Synthia training set and includes fast turning and night scenarios. It was expected that bi3dof's performance on the nuScenes test sets would drop. What was unexpected is that an opposite outcome was observed in methods \cite{cai:iccps2020, shreyas:emsoft2020wip}. To this end, we suspect that this could be a benefit induced by color contrast, i.e., white rain is more salient over darker images because methods \cite{cai:iccps2020, shreyas:emsoft2020wip} learn from color images directly. The average intensity of ID images in the nuScenes test set is 94.8 out of 256, while it is 116.9 in the Synthia test set.  

For applications which require detection, a decision threshold is required. Having a 95\% TPR for an OoD detector is acceptable, i.e., tolerating a 5\% miss rate; the performances of each detector are shown in Table \ref{table:2}. We can see that on nuScenes test sets, bi3dof outperforms \cite{cai:iccps2020} in both F1 score and precision, and \cite{cai:iccps2020} outperforms \cite{shreyas:emsoft2020wip}. On Synthia test sets, bi3dof scores the best in each of the OoD classes, in both F1 score and precision. Performance of \cite{cai:iccps2020} is slightly better than \cite{shreyas:emsoft2020wip} except for the sim.motion OoD class.  

The comparison study shows our spatiotemporal-based feature abstraction approach through 3D optical flow provides superior robustness to motion OoD detection. It also reveals that the sim.motion class is the most difficult OoD class to detect in our experiments. Although bi3dof performs the best among the three for this OoD class at 95\% TPR, the precision of bi3dof is still below 0.5 over Synthia test sets. This indicates that a high percentage of ID samples were falsely classified as OoD.  

\begin{table}
  \caption{ Performance of OoD Detectors at 95\% TPR} \label{table:2}
  %\vspace{-1mm}
  \begin{tabular}{r c ccc c c} 
  \hline  
  & \vline & \multicolumn{3}{c}{Synthia} & \vline & nuScenes \\
  Detector & \vline & sim.motion & sim.rain & comp.rain  & \vline & comp.rain  \\
  \hline 
  \multicolumn{7}{c}{F1 Score} \\
  \hline 
  \cite{cai:iccps2020} & \vline & 0.560 & 0.760 & 0.666 & \vline & 0.775  \\
  \cite{shreyas:emsoft2020wip} & \vline & 0.576 & 0.735 & 0.647 & \vline &  0.659 \\
  bi3dof & \vline & \textbf{0.644} & \textbf{0.961}  & \textbf{0.969}  & \vline & \textbf{0.819}  \\
  \hline 
  \multicolumn{7}{c}{Precision} \\
  \hline 
  \cite{cai:iccps2020} & \vline & 0.397 & 0.633 & 0.512  & \vline & 0.653  \\
  \cite{shreyas:emsoft2020wip} & \vline & 0.413 & 0.600 & 0.504  & \vline & 0.504  \\
  bi3dof & \vline & \textbf{0.483} & \textbf{0.961}  & \textbf{0.969}  & \vline & \textbf{0.713}  \\
  \hline 
\end{tabular}
%\vspace{-4mm}
\end{table}

\subsubsection{\textbf{Improving bi3dof with optimal latent prior}} \label{section:4.2.2}
The bi3dof models in the comparison study were trained with objective given in Equations \ref{eq:3} and \ref{eq:4}, i.e., the simple latent prior$\mathcal{N}(0, \mathbf{I})$ was used for model's training and inference. Using the exact same training sets and hyperparameters, we trained bi3dof-optprior models with objective given in Equations \ref{eq:5} and \ref{eq:6}. As proposed in Section \ref{section:3.2}, we obtained the optimal prior $p(z)$ from a random partition of model's training set. The optimal priors used for training bi3dof-optprior models are $\mathcal{N}(0, 0.01\mathbf{I})$ and $\mathcal{N}(0, 0.005\mathbf{I})$ for the Synthia set, in the horizontal and vertical sub-spaces respectively. For the nuScenes set, the horizontal is $\mathcal{N}(0, 0.075\mathbf{I})$ and the vertical is $\mathcal{N}(0, 0.04\mathbf{I})$.

\begin{figure}
  \centering
  \includegraphics[width=1\linewidth]{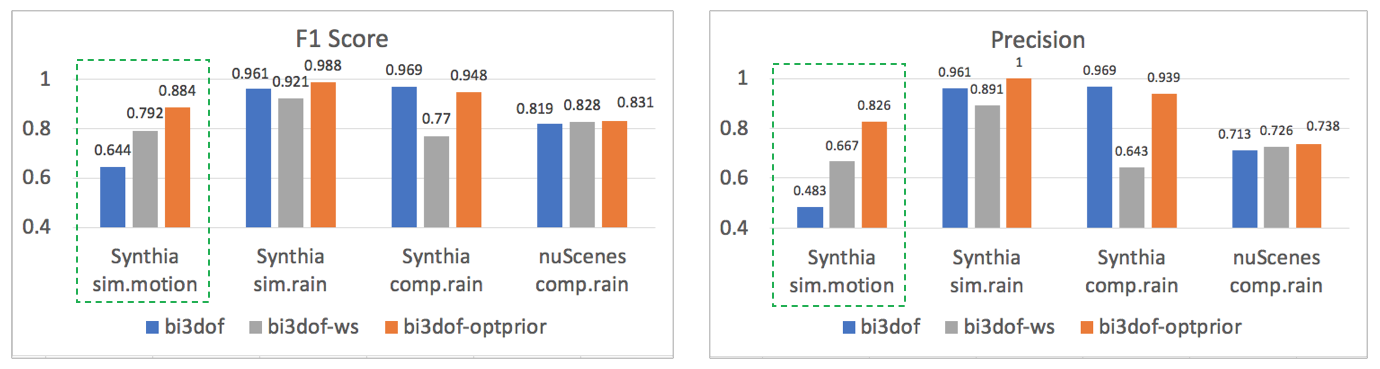}
  \vspace{-4mm}
  \caption{Performance comparison of bi3dof-optprior, bi3dof-ws and bi3dof OoD Detectors at 95\% TPR}
  \label{fig:fig7}
  \vspace{-4mm}
\end{figure}

In bi3dof-optprior, the latent space training is regularized by the 2-order Wasserstein distance instead of KL-divergence. To discern the source of impact to detection performance, we trained a third model regularized with the 2-order Wasserstein but the simple prior. This third model was named bi3dof-ws, represented by grey bars in Figure \ref{fig:fig7}.

At a 95\% TPR, the F1 score and precision of bi3dof-ws does not improve as shown in Figure \ref{fig:fig7}, except for the sim.motion class in the Synthia test set. With the optimal prior, the F1 score and precision further improved from 0.792 to 0.884 and from 0.667 to 0.826, respectively.  In the same figure, we can see a very big drop of F1 score over the comp.rain class for Synthia, from 0.969 to 0.770, in the bi3dof-ws detector. However, with the optimal prior, its F1 score is restored to 0.948.

This group of experiments shows that our optimal prior proposal addresses bi3dof's high false alarm rate over the sim.motion class pointed out in the comparison study, with negligible impact to the detection capacity for other motion OoD classes.

\subsubsection{\textbf{Detection from latent sub-space}} \label{section:4.2.3} 
In our VAE network, horizontal and vertical motion patterns are encoded into two separate latent sub-spaces. When representative OoD samples in both motion directions are available, detection thresholds in each motion direction can be sought separately. We used the rain OoD class as an example to show its benefit. As rain motions largely are in the vertical direction, We took $Score_v$ in Equation \ref{eq:8} to detect rain OoD from ID samples. As shown in Table \ref{table:3}, we can see an overall detection performance improvement when AUROC values from a combined latent space was not close to 100\% correct, while the improvement on those that were already very high is minimal. Unless otherwise specified, we report all experiment results outside this sub-section using OoD scores defined in Equation \ref{eq:8}.

\begin{table}
    \caption{Detection Performance in Latent Sub-space in AUROC } \label{table:3}
    \begin{minipage}{.5\linewidth}
      %\vspace{-2mm}
      \centering
        \begin{tabular}{r c cc c c}
        \hline
        & \vline & \multicolumn{2}{c}{Synthia} & \vline & nuScenes \\        
        \makecell[c]{bi3dof-\\optprior} & \vline  & \makecell[c]{sim.\\rain} & \makecell[c]{comp.\\rain}  & \vline & \makecell[c]{comp.\\rain} \\
        \hline
        $v.$ & \vline  & 0.998 & 0.978 & \vline & \textbf{0.901} \\
        $h.+v.$ & \vline & 1.000 & 0.977  & \vline &  0.877 \\        
        \hline
        \end{tabular}
    \end{minipage}%
    \begin{minipage}{.5\linewidth}
        %\vspace{-2mm}
        \centering        
        \begin{tabular}{r c cc c c}
        \hline
        & \vline & \multicolumn{2}{c}{Synthia} & \vline & nuScenes \\        
        bi3dof& \vline  & \makecell[c]{sim.\\rain} & \makecell[c]{comp.\\rain}  & \vline & \makecell[c]{comp.\\rain}  \\
        \hline
        $v.$ & \vline  & 0.991  & 0.983 & \vline &  \textbf{0.875} \\
        $h.+v.$ & \vline & 0.992 & 0.980 & \vline & 0.839  \\           
        \hline        
        \end{tabular}
    \end{minipage} 
%\vspace{-4mm}
\end{table}

\subsection{Model generalization}  \label{section:4.3}
We believe that without an input space abstraction, the semantic meanings of environmental background will be encoded into the latent space, which discourages a model from generalizing to driving environments that differ from training. However, as pointed out in previous sections, related works only reported their  detection performance on test sets partitioned from the same data set that was used for training. 

In this section, we examined how the models in Section \ref{section:4.2.1} behave if test sets are from a source different from a model's training set. We denote such tests as cross-sets. An example of a cross-set test is to have the nuScenes test set evaluated against a model trained on Synthia data.

\begin{figure}
  \centering
  \includegraphics[width=1\linewidth]{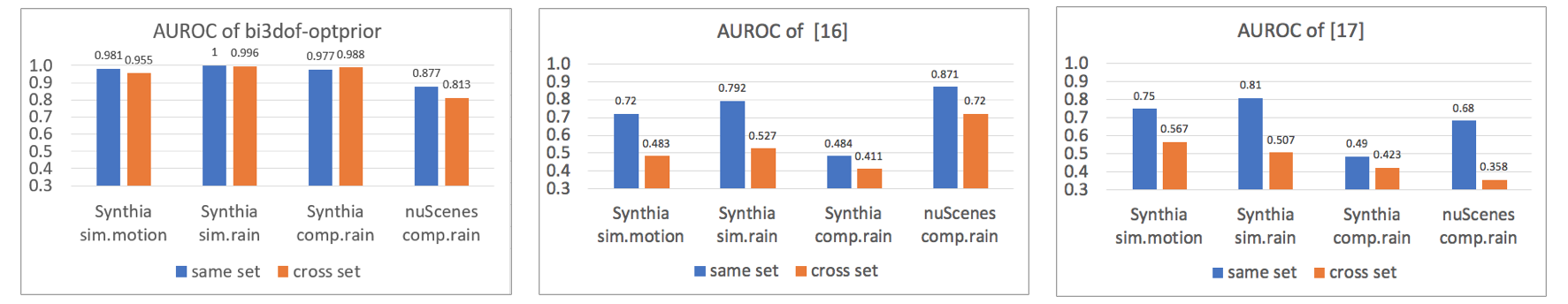}
  %\vspace{-6mm}
  \caption{ Comparison of Model Generalization Capabilities  }
  \label{fig:fig.8}

\end{figure}

From Figure \ref{fig:fig.8}, we can see bi3dof models trained on the nuScenes set generalize perfectly on Synthia test sets across all OoD classes. As the motion ID patterns in the nuScenes training set are more complex than those in the Synthia set, we even observed a trivial performance increase for the Synthia comp.rain class. The maximum performance drop is 0.026 over the sim.motion class. Models trained on the Synthia set generalize slightly less perfectly, having a drop between 0.053 and 0.064. This observation confirms that for a robust ML model, it is vital that an application's population data is well sampled into the model's training set.

While models trained with bi3dof methods generalize very robustly, we can see models from both the related works failing badly.  Six out of eight cross-set tests produced AUROC values around or below 0.5. An AUROC value less than 0.5 indicates a classifier's performance is worse than a random guess, suggesting that the score over ID samples may be statistically higher than OoD samples in most of the cross-set tests here. A relatively good outcome was observed in \cite{cai:iccps2020} on the nuScenes cross-set test, where the AUROC value decreases by 0.151.

\subsection{\textbf{Impact of Input Temporal Window Length}} \label{section:4.4} 
Besides the latent sub-spaces, another novelty in our detection method is enhancing OoD detection's robustness through directly learning 3D optical flow with a sufficiently long temporal window. If a vehicle is moving at 50 km per hour and the video is captured at 25 frames per second, a depth value of 6 encompass motions spanning over 3.3 meters in 240 milliseconds. If the depth value is 1, each optical flow input data point would covers only motions over half a meter. Intuitively, it is less effective in exploiting the motions' correlation over time. Using the nuScenes training set and $depth=1$, we trained a model with optimal prior (i.e., bi2dof-optprior) to validate the effectiveness of learning from spatiotemporal data points. The result in Table \ref{table:4} shows, in both same and cross-set tests, 3D optical flow at depth 6 clearly improve detection performance by a big margin. Specifically, precision values of sim.rain and sim.motion OoD classes in cross-set tests almost double, which indicates false alarm rate dropped dramatically. The 2D optical flow requires less computing capacity, which is an advantage in resource-limited situations.

\begin{table}
  \caption{ Impact of Temporal Window Length on OoD Detection Performance } \label{table:4}
  %\vspace{-1mm}
  \begin{tabular}{r c c c ccc} 
  \hline  
  & \vline  & nuScenes  & \vline & \multicolumn{3}{c}{ Synthia (cross-set) }   \\
  Detector & \vline & comp.rain & \vline  & comp.rain & sim.rain & sim.motion  \\
  \hline 
  \multicolumn{7}{c}{F1 Score} \\ 
  \hline 
  \textbf{bi2dof-optprior} & \vline & 0.698 &  \vline & 0.706 & 0.574 & 0.537   \\
  bi3dof-optprior & \vline & 0.831 &  \vline & 0.959 & 0.976 & 0.864    \\
  \hline 
  \multicolumn{7}{c}{Precision} \\
  \hline 
  \textbf{bi2dof-optprior} & \vline & 0.554 &  \vline & 0.561  & 0.411 & 0.372 \\
  bi3dof-optprior & \vline & 0.738 &  \vline & 0.949 & 1.000 & 0.792  \\
  \hline  
\end{tabular}
\end{table}

\begin{table} 
  \caption{ Entropy Compensated OoD Metric - Detection Performance at 95\% TPR} \label{table:5}
  %\vspace{-1mm}
  \begin{tabular}{r c cccc c ccc} 
  \hline  
  & \vline & \multicolumn{4}{c}{Synthia} & \vline & \multicolumn{3}{c}{nuScenes} \\
  Detector & \vline & \makecell[c]{comp.\\rain} & \textbf{\makecell[c]{comp.\\fog}} & \textbf{darkness} & \makecell[c]{micro-\\average} & \vline & \makecell[c]{micro-\\average} & \textbf{\makecell[c]{comp.\\fog}} & \makecell[c]{comp.\\rain}  \\
  \hline 
  \multicolumn{10}{c}{F1 Score} \\
  \hline 
  \cite{cai:iccps2020} & \vline & 0.666 & 0.689 & 0.690 & 0.684 & \vline & 0.764 & 0.761 & 0.775 \\
  our & \vline  & 0.656 & 0.682 & \textbf{0.810} & \textbf{0.742} & \vline & \textbf{0.803} & \textbf{0.861} & 0.771 \\
  \hline 
  \multicolumn{10}{c}{Precision} \\
  \hline 
  \cite{cai:iccps2020} & \vline & 0.512 & 0.540 & 0.542 & 0.535 & \vline & 0.640 & 0.634 & 0.653  \\
    our & \vline & 0.500 & 0.531 & \textbf{0.706} & \textbf{0.608} & \vline & \textbf{0.695} & \textbf{0.788} & 0.649 \\
  \hline 
  \end{tabular} 
\end{table} 

\subsection{Entropy compensated OoD metric} \label{section:4.5}
So far, our analysis focused on motion OoD detection in the latent space of VAE. This section discusses experimental outcome of enhanced input-space-based detection proposed in Section \ref{section:3.3}. We used \cite{cai:iccps2020}'s method as a baseline to compare OoD classes' performances in the visibility group as given in Table \ref{table:1}. 

Following the definition in Equation \ref{eq:7}, each OoD score is compensated by mean image entropy ratio between the model's training set $entropy(M)$ and the individual test sample. We used VAE models from Section \ref{section:4.2.1} to compute a raw OoD score $MSE(x, \hat{x})$ of a test sample. The baseline entropy, $entropy(M)$, for Synthia and nuScenes models are 4.657 and 5.364, respectively. The difference in $entropy(M)$ values indicates images in the nuScenes data set are averagely more complex than images in the Synthia data set.

Since there are night scenes in the nuScenes training set, darkness detection cannot be tested on nuScenes as dark scenes are considered ID. From Table \ref{table:5}, we can see the entropy compensated OoD metric improves average detection capacity in both Synthia and nuScenes data sets, with negligible impact on the motion OoD classes. Higher precision values over the comp.fog class in nuScenes and darkness class in Synthia indicate that the false alarm rate was reduced. However, the new metric does not improve the F1 score and precision over the Synthia comp-fog class. A pair of comp.fog images in Figure \ref{fig:fig5}. We think this could be a simple mean entropy (in Equation \ref{eq:7}) is not effective for very bright images. A statistically more sophisticated entropy score might worth looking into in future work.

\subsection{Aspects of Network Architecture} \label{section:4.6} 
\begin{figure}[h]
  \centering
  \includegraphics[width=1\linewidth]{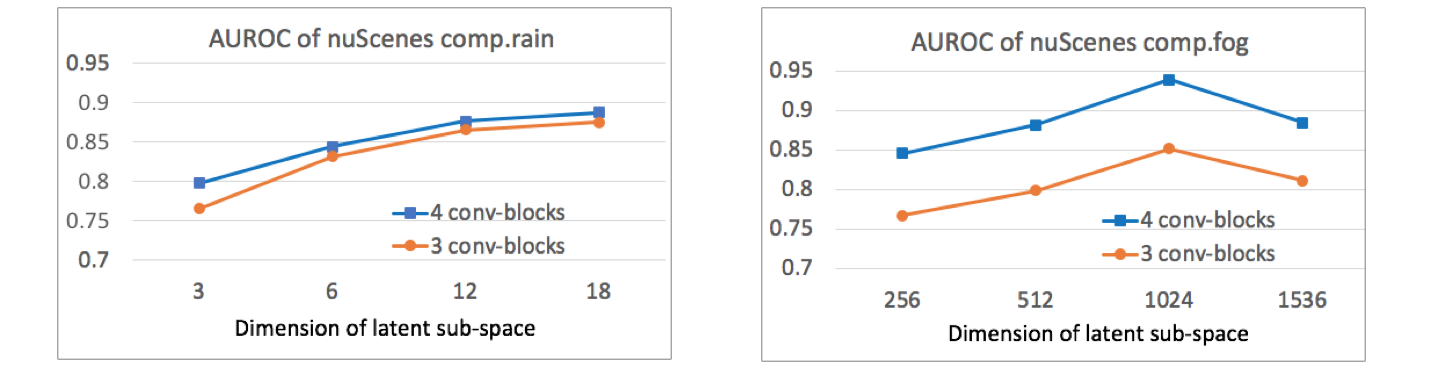}
  \caption{ Influence of Network Architecture to OoD Detection Performance  }
  \begin{small}
     Left is bi3dof-optprior detector. Right is our entropy compensated method from Section \ref{section:4.5}. 
  \end{small}
  \label{fig:fig.9}
\end{figure}

In our comparison studies, all encoders and decoders have four conv-blocks, where each conv-block comprises of one convolutional neural layer, one ELU activation layer, and one BatchNorm layer. In this section, we investigate whether this network choice is optimal for our light-weighted spatiotemporal and entropy-compensated OoD detectors. We experimented with narrower or wider latent dimensions and reduced conv-blocks. The results is summarized in Figure \ref{fig:fig.9}. For the bi3dof-optprior detection method, we can see increasing the dimension of latent space leads to higher performance. When the size of latent space is the same, a network of four conv-blocks consistently outperforms the network with three conv-blocks. Based on the performance curve, four conv-blocks with 12 dimensions for latent sub-space is a good design choice that balances between performance and computational cost. A choice of three conv-blocks with 18 dimensions for latent sub-space is reasonable if one trades performance for less computational cost.  For the entropy-compensated OoD detection method in VAE input space, one more conv-block contributes significantly to detection performance. However, when the latent space dimension is too high, detection performance drops. This can be understood as the information bottleneck required to distill proper abstract representation from underlying data was not formed. A network of four conv-blocks with 1024 dimensions is the optimal choice among all the network parameter space searched. 

\vspace{-4mm}
  
\begin{table}[h]
  \caption{ Computational Cost on a Workstation} \label{table:6}
  %\vspace{-2mm}
  \begin{tabular}{r c ccc c c} 
  \hline  
  & \vline & \multicolumn{3}{c}{Execution Times (milliseconds)} & \vline & Memory size \\
  Detector & \vline & Pre-process & Model inference & Total & \vline &  (MB)  \\   
  \hline 
  \cite{cai:iccps2020} & \vline & 0.2 & 12.5 & 12.7 & \vline & 47.2 \\    
  \cite{shreyas:emsoft2020wip} & \vline & 0.2 & 6.9 & 7.2 & \vline & 14.3 \\  
  bi3dof & \vline & 3.7 & 3.9 & 7.6 & \vline & 8.8 \\   
  \hline   
  \end{tabular} 
  \vspace{-4mm}
\end{table} 

\subsection{Computational cost} \label{section:4.7}

The execution time of an detector matters for OoD detection in real-time,  as well as its memory footprint. The average execution times of each detector on a workstation with an Intel Xeon 3.50GHz is shown in Table \ref{table:6}. The pre-process computation includes image resizing and optical flow operations in the case of bi3dof. The VAE model inference in \cite{cai:iccps2020} takes the longest execution time because the decoder computation is required.  For the same reason, \cite{cai:iccps2020}'s memory footprint is the largest. Our method has the lowest number of dimensions in the latent space, hence the smallest memory footprint. Its execution time is equivalent to \cite{shreyas:emsoft2020wip}. With one execution below 10 milliseconds, all three detection methods can efficiently do real-time OoD detection on general purpose desktop computers. If the neighboring depth-6 3D optical flow results cannot be reused, for example, a frame drop happened while streaming the input video, the total execution time will take 26.3 milliseconds.

\section{OoD Detector for Real-Time Applications on Embedded Devices} \label{section:5}

Many ML-enabled CPS research \cite{almasi2020robust} has been using the Duckietown platform since its launch.  The platform runs on the Robot Operating System (ROS) software system and off-the-shelf mini-computers such as Raspberry Pi. It is also widely used in Internet of Things (IoT) applications. The relatively weak computing power of the Raspberry Pi creates a resource constraint to a direct deployment DNN based OoD detector onto such hardware platform for real-time OoD detection. For example, the execution time of our bi3dof detector is about half a second on a Rasberry Pi 3B+ model (RPI-3B+), which renders directly deploying the bi3dof detector on a Duckiebot for real-time operation infeasible. We introduce a design that customizes our bi3dof method to be deployable onto Google's Coral Edge TPU ~\cite{cass2019taking}. The Edge TPU is optimized for deep neural networks to run real-time inferences on embedded devices. 

With the recent introduction of the new generation Jetson Nano-powered Duckiebot, we have more deployment options. The Jetson Nano is a system-on-chip (SoC) platform. Its CPU clock speed is similar to Raspberry Pi, and it supports model inference in floating-point at 472 GFLOPs. In contrast, the Coral Edge TPU supports integer inference at 4 TOPs but uses USB as a data communication interface. Each embedded platform has its own advantages and disadvantages. For example, we can run model inference directly in floating-point on the Jetson Nano but at approximately one-eighth speed of the Edge TPU. On the other hand, the SoC architecture gives Jetson Nano an advantage in data transmission between the CPU and the model inference unit. A high-end SoC platform, such as Jetson Xavier NX supports model inference in integer number at 21 TOPS. Comparatively, its model inference speed is five times of the Edge TPU. Deploying an integer version of our full-model bi3dof on a high-end embedded platform is thus feasible.     

In the rest of this section, we analyse the computational costs of our model and implementation of \cite{cai:iccps2020, shreyas:emsoft2020wip} on the Raspberry Pi plus Edge TPU embedded platform, introduce a model compression workflow. Finally, we discuss the impact of the model reduction and compression on the OoD detection power. 

\vspace{-1mm}
\subsection{Computational Constraints of Embedded Devices for Real-time Applications} \label{section:5.1}

The RPI-3B+ has a ARMv8 CPU at 1.4Hz. Averagely, one optical flow operation takes 31.7 milliseconds, and one model inference takes 117.7 milliseconds. If all RPI-3B+ computing resource is dedicated to the OoD detector, to achieve zero frame drop in real-time means the video streaming has to be set at 6 frames per second (FPS). In the meantime, the bi3dof model used in Section \ref{section:4} experiments requires 7 consecutive frames from the input video stream to do one OoD detection process. This will incur a long time delay between the image frames captured and when an OoD result is produced, which does not make it suitable for real-time applications on common off-the-shelf embedded devices. 

To adapt to the RPI-3B+ CPU, we used the bi2dof-optprior model trained on the nuScenes training set. After reducing the input space dimension, the model takes about 200 millisecond to produce an OoD result on a RPI-3B+ mini-computer. However, the model inference time of bi2dof is still very long, compared to 7.6-26.3 milliseconds of the bi3dof model running on a workstation. After pushing the model onto the Edge TPU, the VAE model inference time can be reduced on average over four times. In the rest of this section, we discuss the pros and cons of applying model compression, including its impact to the model's detection capability and compare with related works in Section \ref{section:4}. 

\begin{table} 
  \caption{ Computational Cost on a RPI-3B+ Mini-computer} \label{table:7}
  %\vspace{-1mm}
  \begin{tabular}{r c ccc c c} 
  \hline  
  & \vline & \multicolumn{3}{c}{Execution Times (milliseconds)} & \vline & Memory size \\
  Detector & \vline & Pre-process & Model inference & Total & \vline &  (MB)  \\   
  \hline 
  \multicolumn{7}{c}{Floating-point model } \\ 
  \hline
  bi2dof-optprior & \vline & 31.7 & 117.7 & 149.4 & \vline & 8.7 \\   
  \cite{cai:iccps2020}-variant & \vline & 9.9 & 336.6 & 346.5 & \vline & 101.0 \\ 
  \cite{shreyas:emsoft2020wip} & \vline &  9.9 & 220.6 & 230.5 & \vline & 14.3 \\ 
  \hline 
  \multicolumn{7}{c}{8-bit-integer model } \\ 
  \hline
  bi2dof-optprior & \vline & 31.7 & 26.6 & 58.3 & \vline & 2.4 \\   
  \cite{cai:iccps2020}-variant & \vline & 9.9 & 482.1 & 492.0 & \vline & 25.5 \\ 
  \cite{shreyas:emsoft2020wip} & \vline & 9.9  & 10.7 & 20.6 & \vline &  3.7 \\  
  \hline   
  \end{tabular} 
  %\vspace{-4mm}
\end{table}  

\subsection{Model Compression} \label{section:5.2}
DNN models rely on millions or even billions of parameters to achieve superb prediction abilities. For example, the famous AlexNet has over 60 million parameters that require floating point operations. Our floating-point CPU model has 2.2 million parameters. Graphics processing units (GPUs) have been the horsepower for training and inference of DNN models. In recent years, Tensor Processing Units (TPUs), together with various DNN model compression and acceleration techniques, are overcoming challenges in deploying DNN models to embedded devices with limited resources. Parameter quantization is a model compression technique that reduces the number of bits required to store each parameter. With a minimal loss in classification accuracy, train and infer in 8-bit quantized integers, a running time reduction of 50\% was reported on the open-source TensorFlow \cite{jacob2018quantization}.

\begin{figure}[h]
  \centering
  \includegraphics[width=1\linewidth]{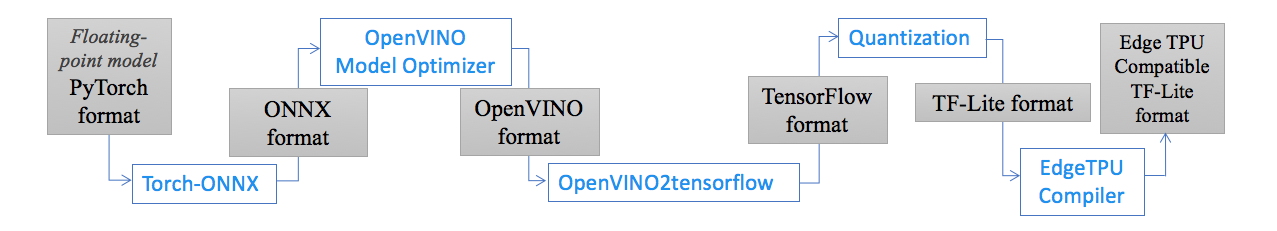}
  \vspace{-6mm}
  \caption{Model Compression Workflow }
   \vspace{1mm}
    \begin{small}
     ONNX provides a toolkit for converting machine learning models built with various frameworks into an open standard format. OpenVINO is a machine learning toolkit from Intel. It was used here to optimize the ONNX format and then convert tensors from channel left (NCHW) to channel right (NHWC) format that TensorFlow supports. Quantization was done with the TFLite-Convertor.
  \end{small}
  \label{fig:fig.10}
\end{figure}

Instead of training a new model in 8-bit directly, we took the 32-bit floating point PyTorch model (bi2dof-optprior) from Section \ref{section:4.5} and applied a full integer post-training quantization. The quantized model is then compiled into the Google Edge TPU's supported format. The conversion workflow is outlined in Figure \ref{fig:fig.10}. Applying the same procedure, we compressed our implementation of \cite{shreyas:emsoft2020wip} in Section \ref{section:4} and a variant implementation of \cite{cai:iccps2020}. Model graphs of all final  outputs and intermediate output of our model are shown in Appendix \ref{appendix:3}. 

A variant implementation is needed because as \cite{cai:iccps2020} does OoD detection in the input space, its decoder is required and needs to be compressed. However, the OpenVINO model optimizer doesn't support the  MaxPool layers in the encoder to pass on pooling indices to the MaxUnPool layers in the decoder. As such, we removed all MaxPool and MaxUnPool layers in the variant implementation. This ended up increasing a significant number of neurons to the linear layer after pooling in the variant model. %while keeping its performance is close to the original \cite{cai:iccps2020} in Section \ref{section:4}.

The results of computational cost are summarised in Table \ref{table:7}. The compressed 8-bit-integer bi2dof-optprior, is 3.6 times smaller and 2.6 times faster than its floating-point model. The compressed 8-bit-integer version of \cite{shreyas:emsoft2020wip} is 3.8 times smaller and 11 times faster than its floating-point model. A lower end-to-end reduction ratio of execution time in bi2dof-optprior is because two model inferences are invoked in the integer version, one for the horizontal and one for the vertical branch, including an  extra pre-process computing (optical flow). The 8-bit-integer version of \cite{cai:iccps2020}-variant is 3.9 times smaller than its floating-point model because most of the decoder's layer computation could not to be pushed onto the Edge TPU, as shown in Appendix \ref{appendix:3}, Figure \ref{fig:fig.a.3-3}. The execution time however, increased by 42\% instead, which implies that the time savings from the computation of the encoder branch on Edge TPU was wiped out by the time spent transmitting the decoder's intermediate result back to the host via USB 2.0. 

\subsection{Impacts of model compression and input-space reduction to performance}\label{section:5.3}

Quantization approximates floating-point values into integers with $ int8\_value = real\_value / scale + zero\_point  $.  It has been reported that its impact on image classification models' performance is minimal. For example, the ImageNet trained Inception\_V4 top-1 accuracy is 80.1\%, as reported in the TensorFlow guideline. Its corresponding quantized model top-1 accuracy is 79.5\%. In classification tasks, the integer vector output can be used directly for class prediction.  

In our case, integer-8 bi2dof-optprior outputs are mean and variance values to be plugged in to Equation \ref{eq:8} to derive an OoD score. We used the scale and zero\_point values shown in Appendix \ref{appendix:3}, Figure \ref{fig:fig.a.3-1} for converting integer values to floating values and vice versa. The process requires a typical data set for calibrating the range of floating-point input tensors. We used bi2dof's training set as the typical data set. a same compression workflow was applied to \cite{shreyas:emsoft2020wip} and \cite{cai:iccps2020}-variant.

\begin{figure}[h]
  \centering
  \includegraphics[width=1\linewidth]{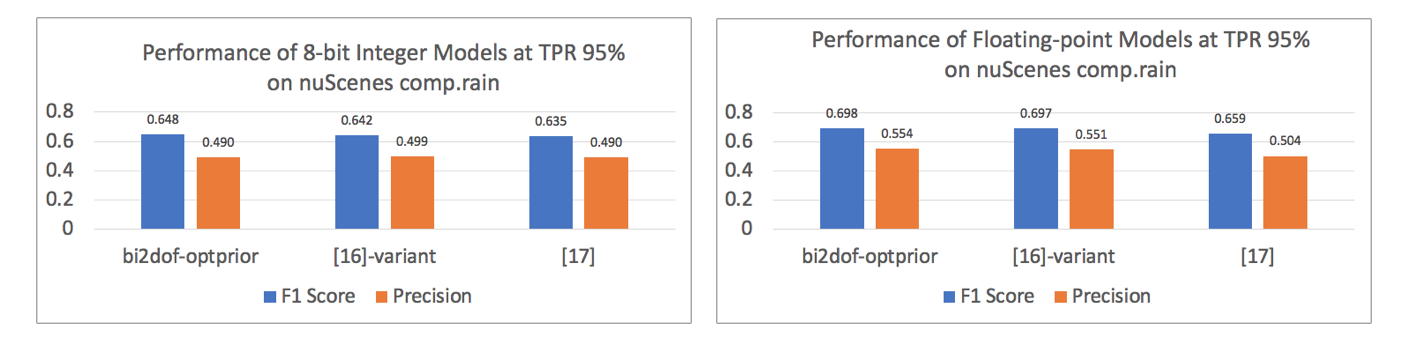}
  \vspace{-4mm}
  \caption{ Impact of Model Compression on OoD Detection Capacity }
  \label{fig:fig.11}
  \vspace{-2mm}
\end{figure}

Comparing detection performance of 8-integer models in Figure \ref{fig:fig.11} with corresponding floating-point versions, we can see that the quantized model's performance dropped a little across all three. While precision is tired up, the F1 score of integer-8 bi2dof-optprior remains higher than the rest. Performance-wise the 8-integer \cite{cai:iccps2020}-variant is closer to our method, but it is not deployable on the RPI + Edge TPU platform. Because its end-to-end execution time is 492 milliseconds.

The result from integer-8 bi2dof-optprior is comparable with its corresponding 32-bit floating-point model, although its performance loss is not minimal as compared to well-known classification tasks. The conversion of integer outputs to float values for OoD score calculation could be one reason for losing more precision in our case. 

\subsection{Section Summary}\label{section:5.4} 
We explored the resource challenges of deploying a DNN-based OoD detector onto off-the-shelf embedded hardware platforms for real-time detection. It showed a compressed version of our detector completed one detection at 58.3 milliseconds at an accuracy comparable with its corresponding floating-point model. That is, given an embedded hardware platform with RPI-3B+ and Google Edge TPU dedicated for OoD detection, our compressed integer-8 bi2dof-optprior, can do one execution without any frame-drop when streaming real-time video at 17 FPS. On the other hand, we showed that reducing input data point in the temporal dimension caused the OoD detection to be less robust.  One optical flow operation on the aforementioned CPU takes 31.7 milliseconds (see in Table \ref{table:7}).  Handling input data points with a larger temporal dimension on off-the-shelf embedded hardware platforms such as Rasberry Pi is challenging. One option for future implementation and deployment of our OoD detector onto a similar embedded platform is to design a FIFO buffer to cache the previous optical flow computation results for a bi3dof OoD detector to reuse. In this feasibility study, we established a bottom line on performance and computational cost.  

\section{CONCLUSION and FUTURE WORK}\label{section:6}
This paper proposed a robust OoD detection method through input space feature abstraction and exploiting the spatiotemporal correlation of motion in videos. Tapping into the prior knowledge in data, we also enhanced the existing VAE-based OoD detection method. Through a series of experiments, we validated the novelties of our new detection methods and enhancement proposals. We showed that our methods' detection performance is significantly better than the state-of-the-art, especially on model generalization capabilities. We plan to deploy our algorithm onto off-the-shelf embedded devices for OoD detection in real-time in future work. In this paper, we compressed our detection model to suit a mini-computer's computing capacity and a recently available Edge TPU. Finally, we provided a feasibility analysis on deployment to an embedded platform, impacts of model compression on detection performance, and how its execution latency affects real-time applications.

%% ACKNOWLEDGE 
\begin{acks}
This research was funded in part by MoE, Singapore, Tier-2 grant number MOE2019-T2-2-040.
\end{acks}

%% REFERENCE
\printbibliography

%% APPENDICES
\clearpage
\begin{appendices}

\section{Likelihood Overlap Explained}\label{appendix:1} 
We trained two VAE models to reproduce the likelihood overlap problem. The mean square errors between the input and reconstructed images were used as OoD scores.  The model encoder includes three CNN blocks of 32/64/128 x (3/4/5) filters with ReLU activation. The decoder is symmetric. And the latent space dimension is 32.

\begin{figure}[h]
  \centering
  \includegraphics[width=0.9\linewidth]{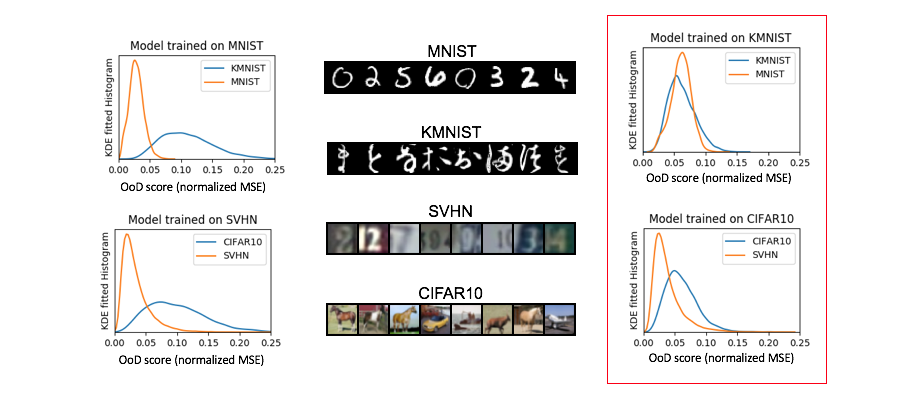}
  \vspace{-4mm}
  \caption{ OoD Detection in Cross Data Domain Setting }
  \begin{small}
Sample images from each data set is shown in the middle column. Distributions of OoD scores from test sets are shown in the left and right columns (likelihood overlap cases) . 
  \end{small}
  \label{fig:fig.a.1}
\end{figure}

 As shown in Figure \ref{fig:fig.a.1}, distributions of OoD scores of KMNIST and MNIST test sets (right column top row) overlap when the detection model is trained with the KMNIST training set. In contrast, when the detection model is trained using the same network but MNIST training set, the  OoD scores distribution of the KMNIST test set (left column top row) clearly diverges from OoD scores of the MNIST test set. The phenomenon is not symmetric with regard to the model training set. This pattern can be observed in CIFAR10 versus SVHN sets as well, in a weaker form.
 
\vspace{-2mm}

\section{ Derivation of Equation \ref{eq:3.2}  } \label{appendix:2}
For abbreviation, A shorter version of notation is used here, i.e. $z$ denotes $z_{h,v}$ in Equation \ref{eq:3.2}, $x$ for $\nabla x_{t}$ and $\hat{x}$ for $\nabla \hat{x_{t}}$.

Let $\hat{x}_{i}$ be the $i^{th}$ value of a Gaussian decoder's reconstruction of $x$ in dimension $M$ and parameters $\mu(x)$ and $\sigma(x)=\mathbf{I}$ to model. The decoder probability density function 
$ p = \frac{1}{\sqrt{2\pi\sigma^{2}(x)}} exp\bigg( -\frac{1}{2} \frac{(\hat{x}_{i}-\mu(x))^2 }{ \sigma^{2}(x) } \bigg) $ 
and the likelihood is $\frac{1}{ (2\pi\sigma^{2}(x))^{M/2}} exp\bigg( -\frac{1}{2\sigma^{2}(x)}  \sum_{i=1}^{M}  (\hat{x}_{i}-\mu(x))^2 \bigg) $. So the negative log-likelihood 

\begin{equation}
\label{eq:9}
\begin{split}
& - \mathbb{E}_{q(z \mid x)} \big[ \log p (x \mid z) \big]  \\
& = \ln \bigg( \big(2\pi\sigma^{2}(x)\big)^{M/2} \bigg) +  \frac{1}{2\sigma^{2}(x)}  \sum_{i=1}^{M}  \big(\hat{x}_{i}-\mu(x)\big)^2 \\
& = M \ln(2\pi) + \frac{1}{2}\sum_{i=1}^{M} (\hat{x}_{i} - x_{i} )^{2} \\
& = const + \frac{M}{2}MSE( \hat{x}, x )  
\end{split}
\end{equation}

\clearpage
\section{Model Graphs from the Compression Workflow} \label{appendix:3}

\begin{figure}[h]
  \centering
  \includegraphics[width=0.54\linewidth]{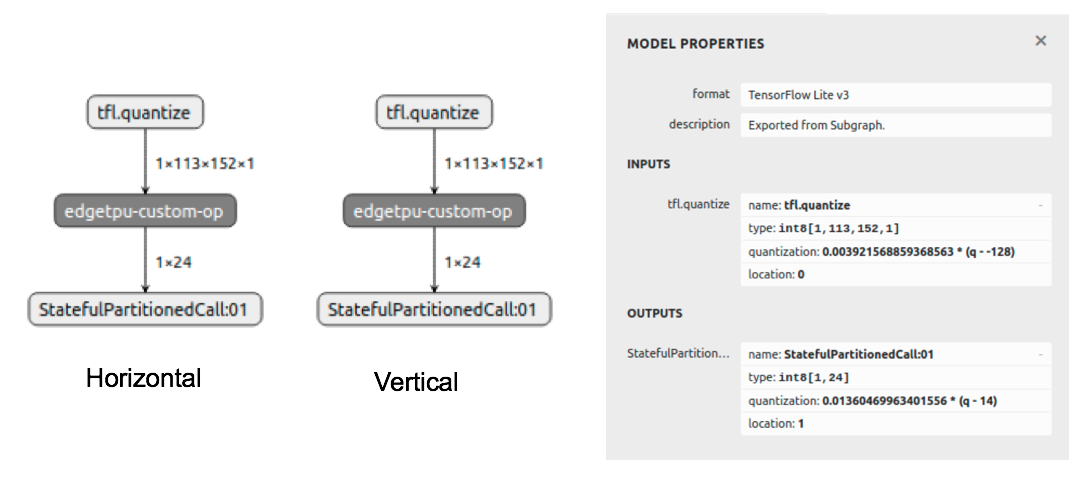}
    %\vspace{-4mm}
  \caption{Model Graph of Twin-Encoder Deployed to Google Edge TPU}
    \begin{small}
    Scale and zero\_point values used for converting input data from floating points to 8-bit integers and output data in the horizontal branch back to floating points are shown in the model properties panel. Scale and zero\_point values of the vertical branch are 0.01613845 and 13, respectively.
  \end{small}
  \label{fig:fig.a.3-1}
\end{figure}

\begin{figure}[h]
  \centering
  \includegraphics[width=0.45\linewidth]{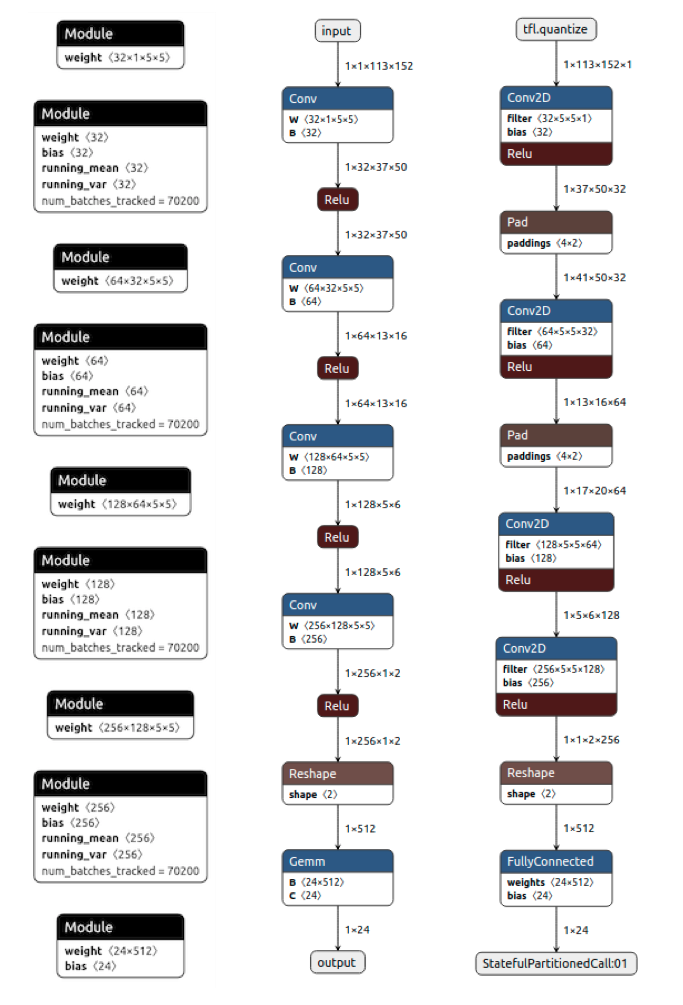}
  \caption{Model Graphs of Intermediate Outputs from the Model Compression Workflow}
  \begin{small}
    Only the horizontal encoder is shown. The vertical encoder network structure is identical. From left to right are the encoder network structure in PyTorch format, ONNX format, and TFLite format after quantization. Activation function ReLU was used here instead of ELU because the Google Edge TPU doesn't support ELU. The OpenVINO tool shifts channel right (NCHW) tensors into channel left (NHWC) format. Its output is in XML format, not shown here. 
  \end{small}
  \label{fig:fig.a.3-2}
\end{figure}

\clearpage
\begin{figure}[h]
  \centering
  \includegraphics[width=0.45\linewidth]{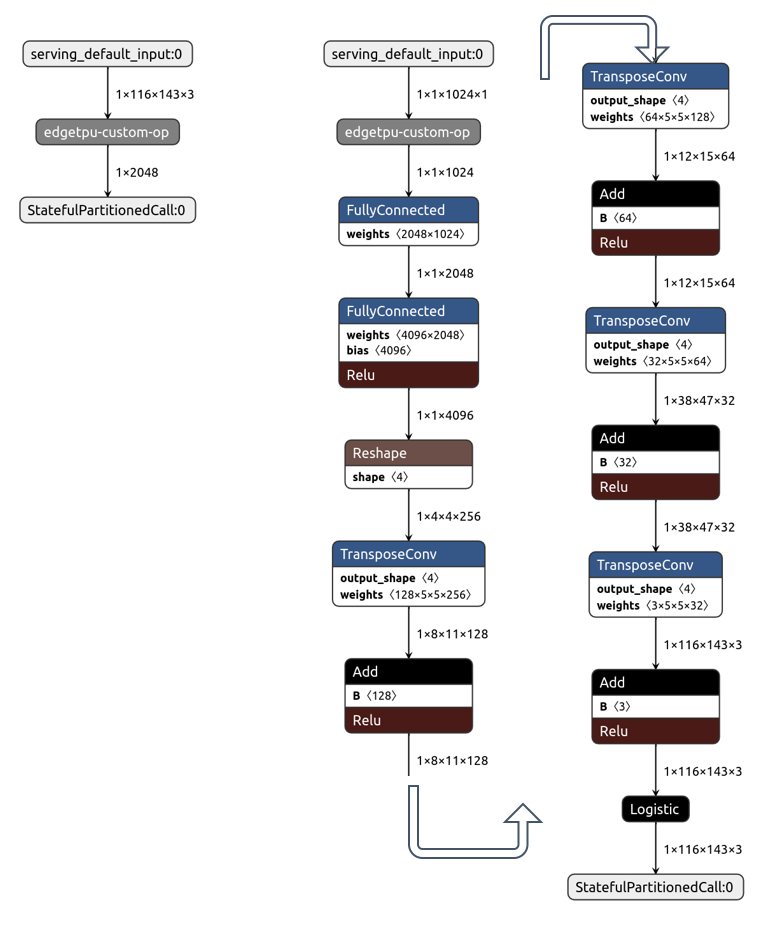}
  \caption{Model Graphs of 8-bit-integer \cite{cai:iccps2020}-variant }
  \begin{small}
  The graph in the left-most column is the encoder. The middle and right columns are graphs of the decoder. Computation of grey-colored layers will be executed on Edge TPU. Computation of none-grey-colored layers will be performed on the CPU.
  \end{small}
  \label{fig:fig.a.3-3}
\end{figure}

\section{Latent Variables Selection Algorithm in \cite{shreyas:emsoft2020wip} }\label{appendix:4}
\begin{table}[h]
  \begin{flushleft}
  \begin{tabular}{cl}
    \toprule
    Pseudo code&\\
    \midrule
    \makecell[r]{Input:} & \makecell[l] { calibration set \( C \), number of latent variables to select \( n \) } \\
    \makecell[r]{Output:} & \makecell[l] {  index of selected latent variables  } \\
    \makecell[r]{1:} & \makecell[l] { \( KL_{diff}=empty \; list \) } \\
    \makecell[r]{2:} & \makecell[l] { For each scene in \(  C  \) do} \\
    \makecell[r]{3:} & \makecell[l] {  \;\;\; \(  F = \) ordered frames belongs to the scene} \\
    \makecell[r]{4:} & \makecell[l] { \;\;\; For  \(  i=0; \; i<len(F); \; i=i+2  \) do} \\
    \makecell[r]{5:} & \makecell[l] { \;\;\;\;\;\; \(  KL^{i} (x_i)=D_{KL} ( q_{\phi}(z_i \mid x_i) \mid N(0,1) \) } \\        
    \makecell[r]{6:} & \makecell[l] { \;\;\;\;\;\; \(  KL^{i+1} (x_{i+1})=D_{KL} ( q_{\phi}(z_{i+1} \mid x_{i+1}) \mid N(0,1) \) } \\      
    \makecell[r]{7:} & \makecell[l] { \;\;\;\;\;\; \(  KL_{diff}^i=\mid KL^{i+1} (x_{i+1}) - KL^{i} (x_i) \mid \) } \\      
    \makecell[r]{8:} & \makecell[l] { \;\;\;\; end for}\\           
    \makecell[r]{9:} & \makecell[l] { end for}\\      
    \makecell[r]{10:} & \makecell[l] { \( AvgKL_{diff}=sort(mean(KL_{diff}, axis=latent\;space\;dimension), inverse=True)  \) }\\      
    \makecell[r]{11:} & \makecell[l] { return index of first \( nth \) latent variables in  \( AvgKL_{diff}  \) }\\                                 
  \bottomrule
\end{tabular}
\end{flushleft}
\end{table}

\end{appendices}

\end{document}